\documentclass[journal,twoside,web]{ieeecolor}
\usepackage{etoolbox}
\makeatletter
\@ifundefined{color@begingroup}%
{\let\color@begingroup\relax
\let\color@endgroup\relax}{}%
\def\fix@ieeecolor@hbox#1{%
\hbox{\color@begingroup#1\color@endgroup}}
\patchcmd\@makecaption{\hbox}{\fix@ieeecolor@hbox}{}{\FAILED}
\patchcmd\@makecaption{\hbox}{\fix@ieeecolor@hbox}{}{\FAILED}
\UseRawInputEncoding
\usepackage{tmi}
\usepackage{cite}
\usepackage{amsmath,amssymb,amsfonts}
\usepackage{algorithmic}
\usepackage{graphicx}
\usepackage{multirow}
\usepackage{textcomp}
\usepackage{bbm}
\usepackage{subfigure} 
\usepackage{booktabs}
\usepackage{multirow}

\usepackage{makecell}

\usepackage{hyperref}
\usepackage[english]{babel}
\usepackage[dvipsnames,svgnames,x11names]{xcolor}

\usepackage[markup=underlined]{changes}

\definechangesauthor[color=blue]{R1}
\definechangesauthor[color=blue]{R2}
\definechangesauthor[color=blue]{R3}
\definechangesauthor[color=blue]{R4}

\hypersetup{
    colorlinks=true,
    linkcolor=cyan,
    filecolor=magenta,      
    urlcolor=cyan,
}

\newcommand{\ETAL}{\textit{et al.}}

\usepackage{amsmath}
\DeclareMathOperator*{\argmax}{arg\,max}
\DeclareMathOperator*{\argmin}{arg\,min}
\DeclareMathOperator*{\onehot}{OneHot}

\def\BibTeX{{\rm B\kern-.05em{\sc i\kern-.025em b}\kern-.08em
    T\kern-.1667em\lower.7ex\hbox{E}\kern-.125emX}}

\begin{document}
\markboth{\journalname, VOL. XX, NO. XX, XXXX 2022}
{Author \MakeLowercase{\textit{et al.}}: Preparation of Papers for IEEE TRANSACTIONS ON MEDICAL IMAGING}

\title{FVP: Fourier Visual Prompting for Source-Free Unsupervised Domain Adaptation of Medical Image Segmentation}

\author{Yan Wang, 
        Jian Cheng,
        Yixin Chen, 
        Shuai Shao,
        Lanyun Zhu, 
        Zhenzhou Wu, 
        Tao Liu, 
        Haogang Zhu
\thanks{This research received support from the National Key Research and Development Program of China under Grant No. 2021ZD0140407, the National Natural Science Foundation of China under Grant No. U21A20523, No. L222152 and No. 61971017 and the Special Project for Innovation in Next-Generation Electronic Information Technology under Grant No.20310105D and the Fundamental Research Funds for the Central Universities.}
\thanks{Yan Wang is with School of Instrumentation and Optoelectronic Engineering, 
and State Key Laboratory of Software Development Environment, Beihang University, Beijing, 100191, China. (Email: wangyan9509@gmail.com).}
\thanks{Shuai Shao is with Zhejiang Lab, Hangzhou, 311100, China.}
\thanks{Lanyun Zhu is with Information Systems Technology and Design (ISTD) pillar, Singapore University of Technology and Design.}
\thanks{Yixin Chen and Zhenzhou Wu are with the BioMind Technology Center, Beijing, China.}
\thanks{Tao Liu is with the School of Biological Science and Medical Engineering, Beihang University, China.}
\thanks{Jian Cheng is with State Key Laboratory of Software Development Environment, Beihang University, Beijing, 100191, China.
(Email: jian\_cheng@buaa.edu.cn).} 
\thanks{Haogang Zhu is with State Key Laboratory of Software Development Environment, Beihang University, Beijing, 100191, China and Zhongguancun Laboratory.
(Email: haogangzhu@buaa.edu.cn).} 
\thanks{$^*$Haogang Zhu and $^*$Jian Cheng are the corresponding authors.}}
\maketitle

\begin{abstract}
Medical image segmentation methods normally perform poorly when there is a domain shift between training and testing data. 
Unsupervised Domain Adaptation (UDA) addresses the domain shift problem by training the model using both labeled data from the source domain and unlabeled data from the target domain. 
Source-Free UDA (SFUDA) was recently proposed for UDA without requiring the source data during the adaptation, due to data privacy or data transmission issues, 
which normally adapts the pre-trained deep model in the testing stage. 
However, in real clinical scenarios of medical image segmentation, the trained model is normally frozen in the testing stage. 
In this paper, we propose Fourier Visual Prompting (FVP) for SFUDA of medical image segmentation. 
Inspired by prompting learning in natural language processing, 
FVP steers the frozen pre-trained model to perform well in the target domain by adding a \textit{visual prompt} to the input target data. 
In FVP, the visual prompt is parameterized using only a small amount of low-frequency learnable parameters in the input frequency space, 
and is learned by minimizing the segmentation loss between the predicted segmentation of the prompted target image and reliable pseudo segmentation label of the target image under the frozen model. 
To our knowledge, FVP is the first work to apply visual prompts to SFUDA for medical image segmentation. 
The proposed FVP is validated using three public datasets, 
and experiments demonstrate that FVP yields better segmentation results, compared with various existing methods.
\end{abstract}


\section{Introduction}
\label{sec:introduction}

\begin{figure}[t]
    \centering
    \includegraphics[width=1.0\linewidth]{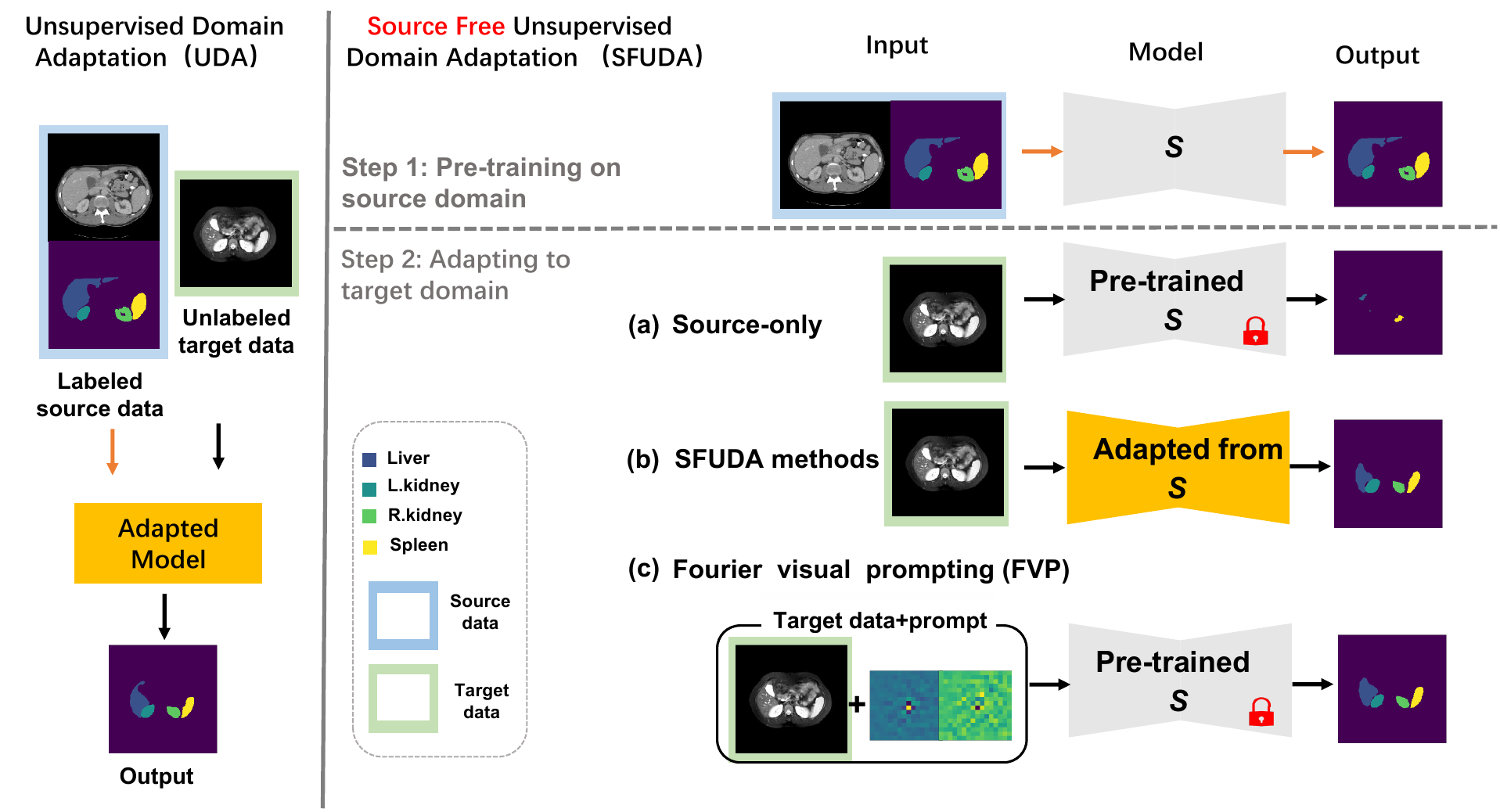}
    \caption{\label{fig:SFUDA}\textbf{An overview of source-free UDA (SFUDA) v.s.\ UDA.} 
    UDA: a deep model is trained based on labeled source data and unlabeled target data. 
    There are two steps in SFUDA: 
    (1) Pre-training on the source domain: training a segmentation network $S$ using the labeled source data (e.g., labeled CT images). 
    (2) Adapting to unlabeled target data (e.g., MR images), without requiring the source data: 
    (a) Source-only: directly inference the target data by the pre-trained model $S$ without adaptation, which results in a performance degradation. 
    (b) SFUDA: perform the adaptation by updating $S$ using unlabeled target data. 
    (c) the proposed Fourier Visual Prompting (FVP): perform the adaptation by updating the target data using a visual prompt in the frequency domain, 
    where the pre-trained model $S$ is frozen, and the visual prompt is learned using unlabeled target data and the frozen $S$ with a self-training strategy.}
\end{figure}

Medical image segmentation plays an important role in clinical applications like computer aided diagnosis. 
In recent years, deep learning methods have been widely used in medical image segmentation~\cite{Unet,falk2019u,chen2017deeplab,chen2023sam,tajbakhsh2020embracing}. 
Deep learning methods are data-driven and are normally based on the assumption that the training data and the testing data follow the same distribution. 
However, deep models normally result in a drastic performance degradation, if there is data distribution discrepancy (called domain shift) between the training and testing data~\cite{guan2021domain,tajbakhsh2020embracing}. 
Domain shift normally happens in real scenarios of medical image analysis~\cite{guan2021domain,tajbakhsh2020embracing}, 
when training and testing medical image data are obtained from different sites, different scanners, same scanners with different scanning parameters, 
or even different modalities, e.g., computed tomography (CT) and magnetic resonance imaging (MRI). 

To address the domain shift issue, unsupervised domain adaptation (UDA) methods are proposed to train the model using both labeled data from the source domain and unlabeled data from the target domain. 
See the left part of Fig.~\ref{fig:SFUDA} for a sketch map of UDA. 
UDA normally relieves the domain shift issue on the image level~\cite{hoffman2018cycada,choi2019self,kang2020pixel}, feature level~\cite{ganin2016domain,chang2019all,SIFA,hong2018conditional}, 
or both the image and feature levels~\cite{chen2020unsupervised}. 
However, UDA still relies on the source data for adaptation, which may be impractical in real scenarios with data privacy or data transmission issues, 
especially for medical images. 

Source-Free UDA (SFUDA) was recently proposed for UDA without requiring the source data during the adaptation~\cite{li2020model,tian2021vdm,SFDA,you2021domain,chen2021source,liang2020we, liu2021adapting}. 
As shown in Fig.~\ref{fig:SFUDA}, SFUDA contains two steps. 
In Step 1, a deep model $S$ is trained using labeled source data. 
In Step 2, the trained model $S$ is updated using unlabeled target data in the testing stage, without requiring the source data. 
Recently, an increasing number of works focus on the challenging SFUDA task. 
Existing SFUDA methods could be divided into two categories: 
GAN-based methods~\cite{li2020model,tian2021vdm,SFDA} and self-training based methods~\cite{you2021domain,chen2021source,liang2020we, liu2021adapting}. 
For example, 3C-GAN~\cite{li2020model} first generates target-style images by conditional GAN, and then collaborates the generator with pre-trained source model for the final adaptation. 
Another research direction is to align features from the source and target domains by self-training strategies using pseudo-labels of target data obtained by the pre-trained model~\cite{you2021domain,EM1,EM2}. 
Existing SFUDA approaches, despite their gratifying results, still rely on updating (e.g., fine-tuning) the pre-trained source model. 
However, in real clinical scenarios of medical image segmentation, the trained model is normally frozen in the testing stage, 
because of software security issues, or Food and Drug Administration (FDA) regulations which may require frozen models for deployment. 
Thus, it limits the clinical usage of existing SFUDA methods, if updating the pre-trained model in SFUDA.

In this paper, we propose Fourier Visual Prompting (FVP) for SFUDA of medical image segmentation, without changing the pre-trained model. 
Inspired by prompting learning in natural language processing, FVP steers the frozen pre-trained model to perform well in the target domain by adding a \textit{visual prompt} to the input target data. 
In FVP, the visual prompt is parameterized using only a small amount of learnable parameters in the input frequency space, 
and is learned by minimizing the segmentation loss between the predicted segmentation of the prompted target image and reliable pseudo segmentation label of the target image under the frozen model. 

The main contributions of this work are summarized as: 
(1) We address a challenging scenario: source-free unsupervised domain adaptation (SFUDA) for medical image segmentation with the frozen pre-trained model, 
which is more suitable for clinical applications.
(2) We propose a novel method called Fourier visual prompting (FVP) for SFUDA. 
FVP steers the frozen pre-trained model to perform well in the target domain by adding a \textit{visual prompt} parameterized by its low frequency components to the input target data.
To our knowledge, FVP is the first work to apply visual prompt to SFUDA for medical image segmentation. 
(3) We design a reliable label detection module for pseudo labels of the target data by the pre-trained model, 
and the detected reliable pseudo labels are used to learn the visual prompt for the unlabeled target data. 
(4) We validate the effectiveness of the proposed FVP by using three public domain adaptation benchmark datasets. 
Experimental results demonstrate that our proposed FVP outperforms several state-of-the-art methods, 
indicating that leveraging the Fourier visual prompt could achieve adaptation in a simple yet effective way.

\section{Related work}

\subsection{Source-Free Unsupervised Domain Adaptation}

Due to the data privacy and data transmission issues in real clinical application scenarios, 
source-free unsupervised domain adaptation (SFUDA) aims to tackle the domain shift without requiring the source data. 
Starting with the definition of SFUDA in~\cite{liang2020we}, it has witnessed an increasing number of SFUDA approaches~\cite{you2021domain, chen2021source, liu2021adapting, huang2021model, qiu2021source, Liang2021source, xia2021adaptive}. 
The existing SFUDA methods often have two categories: GAN-based methods~\cite{li2020model, tian2021vdm, SFDA} and self-training based methods~\cite{you2021domain,chen2021source,liang2020we, liu2021adapting}. 
GAN-based methods aim to restore the source domain distribution through generative adversarial networks (GANs) and then use source-available domain adaptation methods. 
Self-training based methods often consist of two aspects. 
First, align the target domain features or images. 
Second, apply the self-training process and adopt pseudo-labeling. 
Pseudo-labeling aims to mine as much reliable information as possible in the source model's predictions of the target data by various label selection methods. 
For example, \cite{you2021domain} adopts positive and negative learning for correcting the offset of false labels and it actually provides a selection for reliable pseudo labels. 
\cite{chen2021source} solves the pseudo label selections by prototype estimation for the image segmentation task with two classes. 
\cite{liu2021adapting} uses the knowledge of the batch normalization layer to compare the source domain features and target domain features and then achieves domain-wise alignment. 
It is worth noting that all of the mentioned methods still need to update the parameters of the pre-trained model, which may be impractical in some real clinical scenarios. 
On the contrary, our method FVP is to add a learnable visual prompt to the target images under the frozen pre-trained model, 
which could be more suitable and valuable for real clinical scenarios.

\subsection{Prompt Learning}
Prompt learning is first introduced in natural language processing (NLP)~\cite{petroni2019language, prompt_nlp,prompt_tuning,jiang2020can,li2021prefix}. 
Prompting means designing a template to reformulate the downstream dataset, so that the pre-trained frozen model could be directly applied to a new task without updating. 
By manually designing the correct prompt, it can improve the performance of the downstream tasks. 
This idea has been extended to vision tasks combined with vision-language model. 
For example, Tsimpoukelli \ETAL~\cite{prompt_few_shot} train a vision encoder to represent each image as a prefix so that the frozen language model is prompted. 
This multimodal (e.g., vision and language) few-shot learner can learn words for new objects and novel visual categories rapidly. 
Zhou \ETAL~\cite{zhou2021learning} aim to map the input texts and images to the same feature space by contrastive learning, and then design the prompt for the texts. 
Recently, there are few works proposing Visual Prompting only in visual domain \cite{visualprompt_tuning, visualprompt}. 
Visual prompt tuning (VPT)~\cite{visualprompt_tuning} tunes both a learnable visual prompt in the input space and the head of the pre-trained transformer model, 
while keeping the pre-trained model backbone frozen. 
Instead of full fine-tuning of the pre-trained model with a large number of tunable parameters, 
VPT modifies the input images by a learnable visual prompt with only a small amount of tunable parameters in the visual prompt and the head of the pre-trained model. 
Visual prompting~\cite{visualprompt} is also introduced by modifying the pixels of the input images to adapt frozen pre-trained classification models. 
It introduces perturbation in the pixel space of images to improve model performance, 
which is different from adversarial examples~\cite{moosavi2017universal} that slightly change pixels of images to confuse the model. 
A visual prompt is designed to be task and dataset specific, since it is learned from the target domain data for the downstream task. 
The effectiveness of visual prompting in transfer learning provides a new perspective for domain adaptation~\cite{visualprompt_tuning,visualprompt}, 
and the experiments demonstrate that visual prompting achieves better performance than text prompt in image classification task~\cite{visualprompt}. 
However, the visual prompt in the input image space results in mosaic style and noisy prompted input images. 
In this paper, we introduce the Fourier visual prompt (FVP) in the input frequency space, 
so that it can change the input image style in a global manner with less learnable parameters, 
and it also generates visually reasonable prompted images. 

\subsection{Fourier Transform and Spectrum Analysis}

Spectrum analysis by Fourier transform is an important technique in image processing~\cite{bracewell1986fourier,nussbaumer1981fast}. 
%
There are an increasing number of works which incorporate spectrum analysis by Fourier or wavelet transforms into deep neural networks. 
Wang \ETAL~\cite{cvprfourier2020oral} analyze and explain the generalization of Convolutional Neural Networks (CNNs) from the perspective of the frequency domain. 
It proposes the hypothesis that the CNN model first picks up the low frequency component, 
then gradually picks up the high frequency component to achieve higher training accuracy. 

For domain adaptation, recent works~\cite{FDA,yang2020phase,fourierDG} show that 
the amplitude component of the Fourier transform of an image contains more style information, 
while the phase component contains more semantic contents. 
Thus, modifying amplitude component of an image could be used for style transfer as data augmentation. 
FDA in~\cite{FDA} provides a simple method to generate labeled target-like images from source images by replacing the low-frequency amplitude components of the source images with them of the target images, 
then UDA becomes a semi-supervised learning (SSL) problem, since FDA aligns the two domains in the image level. 
Xu \ETAL~\cite{fourierDG} introduce a Fourier-based framework for domain generalization by assuming that the phase component of an image is not easily affected by domain shifts, 
and propose Fourier-based data augmentation by mixing amplitude components. 
Fourier-based domain adaptation methods do not require adversarial optimization to align features from different domains. 
Different from existing works, we apply prompt learning parameterized in the input frequency space to SFUDA, without requiring updating the pre-trained model. 
To the best of our knowledge, the proposed FVP is the first work to apply visual prompt to SFUDA for medical image segmentation. 

\section{Methodology}
\label{sec:methods}
\begin{figure*}[t]
    \centering
    \includegraphics[width=0.9\linewidth]{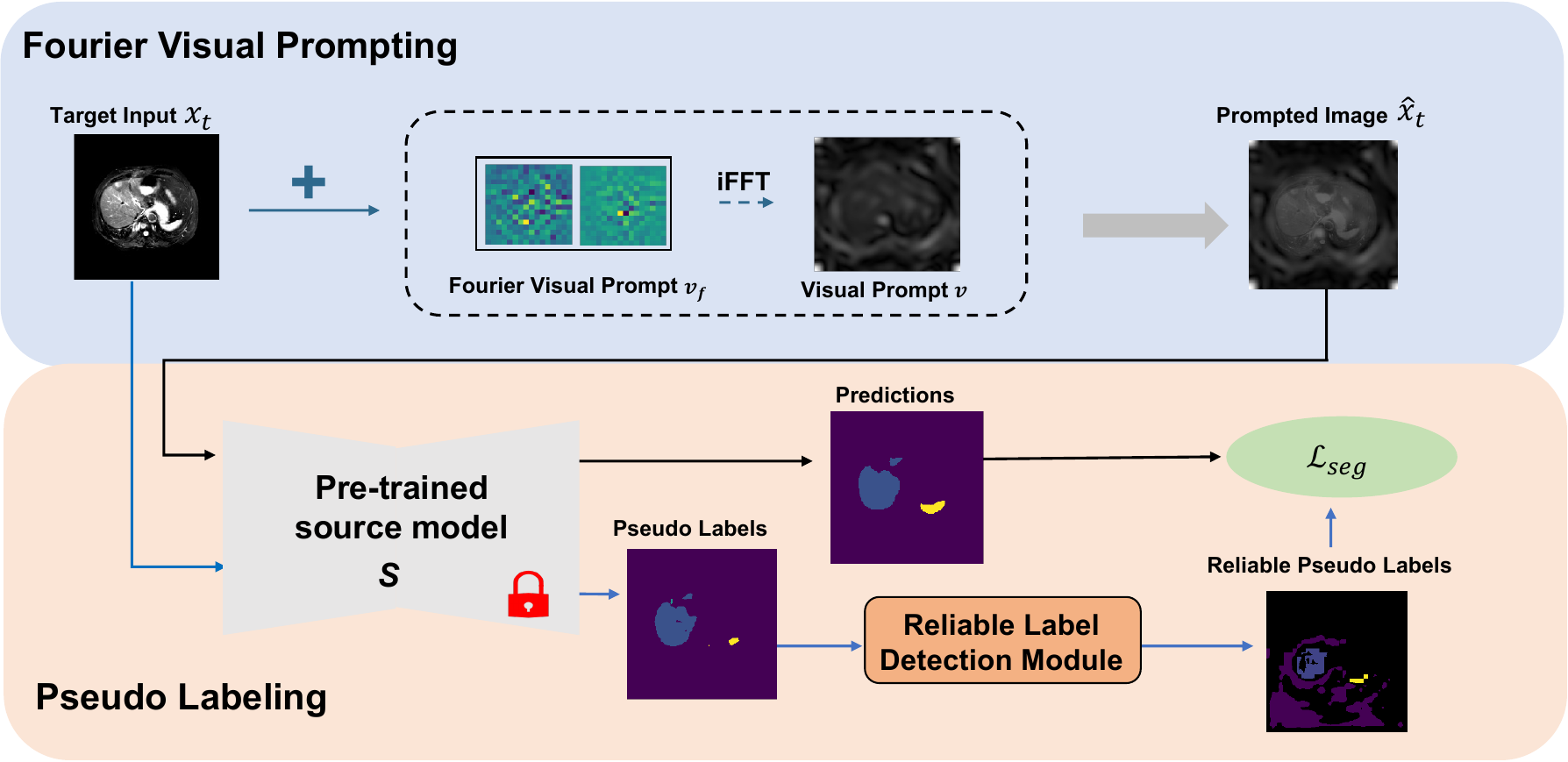}
    \caption{\label{fig:framework}\textbf{The overall framework for training the Fourier Visual Prompt by the self-training process}. 
    Fourier visual prompting: parameterize the visual prompt in the frequency domain and then add the visual prompt after iFFT to the target inputs. 
    Pseudo-labeling: select reliable pseudo labels by Reliable Label Detection Module and then use them to train the Fourier visual prompt.}
\end{figure*}

In this section, we introduce the SFUDA with the frozen pre-trained model in Section~\ref{sec:sfuda}, 
propose the Fourier visual prompt in Section~\ref{sec:FVP}, 
and then propose a pseudo-labeling strategy to train the Fourier visual prompt with a reliable label detection module in Section~\ref{sec:ST}. 
The overall framework is illustrated in Fig.~\ref{fig:framework}. 

\subsection{SFUDA with the Frozen Pre-trained Model}
\label{sec:sfuda}
Denote the source domain dataset with $N_s$ labeled images by $D_s=\{x_{s}^i, y_{s}^i\}_{i=1}^{N_s}$, where $x_{s}^i$ and $y_{s}^i$ denote the $i$-th source image and its label, respectively. 
Denote the target domain dataset with $N_t$ unlabeled images by . 
If the segmentation model $S$ is trained using $D_s$ and it is tested in $D_t$, then it normally results in poor performance due to domain shift, 
as shown in the source-only setting in Fig.~\ref{fig:SFUDA} (a). 
UDA is to train a segmentation model using both datasets $D_s$ and $D_t$, as shown in the left part of Fig.~\ref{fig:SFUDA}. 

Source-free UDA (SFUDA), which does not require source data during adaptation, has two steps in training. 
In the first pre-training step, the segmentation model $S$ is trained only using labeled source domain dataset $D_s$. 
In the second adaptation step, 
existing SFUDA methods (Fig.~\ref{fig:SFUDA} (b)) normally update the pre-trained model $S$ for adaptation~\cite{li2020model,tian2021vdm,SFDA,you2021domain,chen2021source,liang2020we, liu2021adapting}, 
which may be impractical for real clinical scenarios. 
In this paper, we aim to adapt the target images by adding a learnable visual prompt, with the pre-trained model $S$ frozen. 
See Fig.~\ref{fig:SFUDA} (c). 


\subsection{Fourier Visual Prompting}
\label{sec:FVP}

Without loss of generality, we denote a target image $x_t \in \mathbb{R}^{H \times W \times C}$ for the 2D case with spatial size $H\times W$ and channel $C$. 
For a 2D MRI image as an example, $C=1$. 
In order to perform SFUDA with the frozen pre-trained model $S$, 
we propose to learn a \textit{visual prompt} $v \in \mathbb{R}^{H \times W \times C} $ in the voxel space, 
in order to steer the model $S$ to perform well on $D_t$ without updating its parameters. 
The visual prompt $v$ is added to the input image $x_t$ to obtain the prompted image 
\begin{equation}
\hat{x}_t= x_t + v. \label{eq:fvp_v}
\end{equation}
The visual prompt is to introduce perturbation in the voxel space of $x_t$, so that the prompted image $\hat{x}_t$ has better performance than the original image $x_t$.

Different visual prompt designs in the pixel space were tested in~\cite{visualprompt} for transfer learning in image classification,  
including pixel patch in random and fixed locations, and padding. 
Experiments in~\cite{visualprompt} showed that the padding design with less learnable parameters obtains the best performance, 
compared with the full-size visual prompt with total $CHW$ learnable parameters, and pixel patch designs. 
However, the visual prompt of the padding design in the input pixel space results in mosaic style and noisy prompted images with sharp boundary around the padding~\cite{visualprompt}, 
which may not appropriate for the image segmentation task.

In this paper, we propose to parameterize the visual prompt $v$ in the frequency domain of the input image $x_t$, 
so that it can change the input image style in a global manner with less learnable parameters, 
and generate visually reasonable prompted images. 
For a target image $x_t$, its discrete Fourier transform is defined as: 
\begin{equation}
    \mathcal{F}(x_t)(u,q) = \sum_{h=0}^{H-1}\sum_{w=0}^{W-1}x_t(h,w)e^{-(j\frac{2\pi uh}{H}+j \frac{2\pi qw}{W})}, 
\end{equation}
and its amplitude and phase components are $\mathcal{F}^A(x_t)$ and $\mathcal{F}^P(x_t)$, respectively, 
i.e., $\mathcal{F}(x_t)= [\mathcal{F}^A(x_t), \mathcal{F}^P(x_t)]$. 
The inverse Fourier transform is denoted as $\mathcal{F}^{-1}$. 
Both the Fourier transformation and its inverse can be calculated via the Fast Fourier Transform (FFT). 
Then, by considering a complex \textit{Fourier visual prompt} (FVP) $v_f\in \mathbb{R}^{H \times W \times 2C} $ in the frequency domain 
with both amplitude and phase components, 
i.e., $v=\mathcal{F}^{-1}(v_f)$ in Eq.~\eqref{eq:fvp_v},
the prompted image is represented as: 
\begin{equation}\label{eq:fvp_complex} 
  \hat{x}_t= \mathcal{F}^{-1}(\mathcal{F}(x_t) + v_f) = x_t + \mathcal{F}^{-1}(v_f).  
\end{equation}
In order to reduce the actual number of learnable parameters, 
we set high frequency part of $v_f$ as $0$ and only optimize $v_f$ in the low frequency part, 
considering that the low frequency part plays an important role in style transfer and domain adaptation~\cite{FDA,yang2020phase,fourierDG}. 
To be specific, assuming that the zero frequency component of $\mathcal{F}(x_t)$ is in the center, 
we set $v_f$ as $0$ outside a patch mask centered at the frequency image center with a size of $r\times r$, 
i.e, 
\begin{equation}
  v_f(u, q) = 0, \qquad \text{if} \ \ |u|>\frac{r}{2} \text{  or  } |q| >\frac{r}{2}. \label{eq:v_f_box}
\end{equation}

Recent domain adaptation works~\cite{FDA,yang2020phase,fourierDG} 
show that modifying the amplitude component of an image could be used for style transfer as data augmentation, while fixing the phase component. 
In our FVP, we could also add a real $v_f\in \mathbb{R}^{H \times W \times C}$ into the amplitude component of $x_t$, while fixing the phase component, i.e.,  
\begin{equation}
  \hat{x}_t= \mathcal{F}^{-1}( [\mathcal{F}^A(x_t) + v_f, \mathcal{F}^P(x_t)]),  \label{eq:fvp_amplitude}
\end{equation}

or add a real $v_f\in \mathbb{R}^{H \times W \times C}$ into the phase component of $x_t$, while fixing the amplitude component, i.e.,  
$\hat{x}_t= \mathcal{F}^{-1}( [\mathcal{F}^A(x_t) , \mathcal{F}^P(x_t)+ v_f])$. %

It is worth noting that the FVP in Eq.~\eqref{eq:fvp_complex} and Eq.~\eqref{eq:fvp_amplitude} are dataset-specific and input-agnostic, 
which means all the input target images $D_t$ share the same FVP. 
FVP $v_f$ in~\eqref{eq:fvp_complex} is complex, and modifies the image $x_t$ in both the amplitude and phase components. 
While FVP $v_f$ in~\eqref{eq:fvp_amplitude} is real, and modifies the image $x_t$ only in the amplitude component. 
For the complex FVP in~\eqref{eq:fvp_complex}, FFT only needs to be performed once for $v_f$ in each training batch. 
While for the real FVP in~\eqref{eq:fvp_amplitude}, FFT needs to be performed in $2N_b$ times in each training batch if the batch size is $N_b$.  
Thus, the complex FVP in~\eqref{eq:fvp_complex} is much more computationally efficient.  
Moreover, our experiments show that the complex FVP in~\eqref{eq:fvp_complex} yields better segmentation performance. 

Next, our goal is to train the learnable FVP $v_f$ by using the target dataset $D_t$ and the frozen pre-trained segmentation model $S$. 
In this way, we can get the optimized Fourier visual prompt for the target dataset, 
which is input-agnostic across the entire target dataset.

\subsection{Reliable Label Detection Module for Pseudo-labeling}
\label{sec:ST}

In our proposed FVP method, we need to train the learnable FVP $v_f$ by using the unlabeled target data $D_t=\{x_t^i\}_{i=1}^{N_t}$ and the frozen pre-trained model $S$. 
Segmentation predictions directly inferenced by the model $S$ without domain adaptation, i.e., $\{S(x_t^i)\}_{i=1}^{N_t}$, can be seen as pseudo labels. 
Due to the domain shift between source and target domains, these pseudo labels directly inferenced by the pre-trained model are unreliable with noise. 
In this section, we propose a reliable label detection module to select reliable pseudo labels to supervise the learning of the FVP, as shown in Fig.~\ref{fig:framework}. 

\subsubsection{Segmentation Loss Using Pseudo Labels}
Let $p_x = S(x)$ denote the segmentation prediction obtained by the pre-trained model $S$ with the input of $x$. 
Then, for a target image $x_t \in \mathbb{R}^{H \times W \times C}$, its pseudo label is $p_{x_t}=S(x_t) \in [0,1]^{H \times W \times N_c}$, 
where $N_c$ is the number of segmentation classes, and $p_{x_t}^{h,w,c} \in [0,1]$ denotes the output probability for the $c$-th class at the voxel $(h,w)$. 
The output of the prompted image $\hat{x}_t$ is $p_{\hat{x}_t}=S(\hat{x}_t)$, which is a function of the FVP $v_f$. 
Then, the segmentation loss to learn the FVP $v_f$ is proposed as:
\begin{equation}
  \mathcal{L}_{seg} = -\sum_{h,w,c} T^{h,w}\cdot y_{x_t}^{h,w,c}\log p_{\hat{x}_t}^{h,w,c}, \label{eq:loss}
\end{equation}%

where $y_{x_t}$ is a one-hot format reliable pseudo target label generated from the pseudo label $p_{x_t}$ by setting unreliable prediction probabilities to zero and then applying the argmax operator and one-hot encoding, 
and $T$ is a binary selection mask introduced to select reliable pseudo labels on the voxel level, 
$T^{h,w}=1$ means the reliable voxel $(h,w)$ is used in the loss. 
Considering $y_{x_t}^{h,w}$ is a $N_c$-dimensional one-hot vector, and $T^{h,w}$ is binary, 
we could define 
\begin{equation}
R_{x_t}^{h,w,c}=
\begin{cases}
  [y_{x_t}^{h,w,c}; 0], &  \text{if}\ \ T^{h,w}=1 \\
  [\vec{0}_{N_c}; 1], & \text{if}\ \ T^{h,w}=0 
\end{cases}
\end{equation}%
as the final $N_c+1$-dimensional one-hot format reliable pseudo label for voxel $(h,w)$, 
where $\vec{0}_{N_c}$ is the $N_c$-dimensional zero vector. 
$R_{x_t}^{h,w,c}$ encodes the binary $T^{h,w}$ into its last dimension, and encodes $y_{x_t}^{h,w,c}$ into its first $N_c$ dimension. 

Essentially, the segmentation loss in Eq.~\eqref{eq:loss} is a revised cross entropy loss for multi-class segmentation, 
where the binary selection mask $T$ and the one-hot format target label $y_{x_t}$ are introduced 
to select reliable pseudo labels from $p_{x_t}$ on both the voxel level and the prediction probability level, as shown in Fig.~\ref{fig:framework}. 
We will show how to set $T$ and $y_{x_t}$ based on the following double-threshold selection and prototype based selection.


\subsubsection{Double-Threshold Selection}

It is possible to generate one-hot the pseudo target label $y_{x_t}$ by directly applying the argmax operator and one-hot encoding to the pseudo label $p_{x_t}=S(x_t)$, 
which in each voxel assigns the class label based on the highest prediction probability. 
However, this naive pseudo target label generation is prone to result in incorrect labels, due to domain shift and class imbalance.
For class imbalance in segmentation, the naive pseudo target label generation normally results in poor performance in hard-to-transfer classes, 
since it is easy for the model to overfit the majority classes, but ignore the minority classes~\cite{you2021domain}. 
This is a common issue in medical image segmentation, since some organs for segmentation are very small and the class imbalance always happens because of the background class with much more voxels. 
 
For the pseudo label $p_{x_t}$, small prediction probability values are not reliable. 
Therefore, a straightforward strategy is to introduce a global threshold $\lambda$ to select reliable pseudo labels to avoid small probability values. 
In other words, we only consider reliable pseudo label $p_{x_t}^{h,w,c}$ when $p_{x_t}^{h,w,c} \geq \lambda$ at voxel $(h,w)$ in the segmentation loss in Eq.~\eqref{eq:loss}. 

Motivated by previous works on selecting pseudo labels using a threshold~\cite{you2021domain,wang2021give}, 
we introduce an intra-class level threshold to select the voxels with intra-class confidence. 
To avoid the imbalanced selection, the intra-class threshold is defined as: 
\begin{equation}
  \delta_{x_t}^{c} = \tau_{k}(p_{x_t}^{c}), \quad p_{x_t}^c=\{p_{x_t}^{h,w,c} \ | \ \forall  h, \forall w \}, 
\end{equation}
where $p_{x_t}^{c}$ is the output prediction image $p_{x_t}$ at the $c$-th channel, 
and $\tau_k(\cdot)$ denotes the top $k$ value operator. 
Then, for each voxel $(h,w)$ at the $c$-th channel, we would like to only consider the probability prediction $p_{x_t}^{h,w,c}$ whose value is larger than threshold $\delta_{x_t}^{c}$ in the segmentation loss in Eq.~\eqref{eq:loss}, 
and set other values to zero. 

To this end, the selected probability $\hat{p}_{x_t}^{h,w,c}$ under the above two thresholds is:
\begin{equation}
  \hat{p}_{x_t}^{h,w,c}= p_{x_t}^{h,w,c} \cdot \mathbbm{1}(p_{x_t}^{h,w,c} \geq \delta_{x_t}^{c}) \mathbbm{1}(p_{x_t}^{h,w,c} \geq \lambda), \label{eq:threshold}
\end{equation}
where $\mathbbm{1}(\cdot)$ is the indicator function, whose value is $1$ when the condition is satisfied, and $0$ otherwise. 
Essentially, in Eq.~\eqref{eq:threshold}, these two thresholds $\lambda$ and $\delta_{x_t}^c$ are used to set unreliable probabilities to zero. 
To be specific, the revised probability output $\hat{p}_{x_t}^{h,w,c}$ sets the pseudo label probability $p_{x_t}^{h,w,c}$ to $0$, 
if its value is globally small (i.e., $p_{x_t}^{h,w,c} < \lambda$) or if its value is smaller than the intra-class threshold (i.e., $p_{x_t}^{h,w,c} < \delta_{x_t}^c$).

Then, the pseudo target label $y_{x_t}$ in Eq.~\eqref{eq:loss} is generated by applying the argmax operator and one-hot encoding to the revised pseudo label $\hat{p}_{x_t}^{h,w,c}$, i.e., 
\begin{equation}
  y_{x_t}^{h,w} = \onehot(\argmax(\hat{p}_{x_t}^{h,w}), N_c), \label{eq:y} 
\end{equation}
where $\hat{p}_{x_t}^{h,w}$ is the $N_c$ dimensional revised probability vector of $\hat{p}_{x_t}$ at voxel $(h,w)$. 
{Note that when the prediction $p_{x_t}^{h,w,c}$ in Eq.~\eqref{eq:threshold} is smaller than both thresholds $\delta_{x_t}^{c}$ and $\lambda$ for all classes, 
then $\hat{p}_{x_t}^{h,w}$ is a zero vector, and $y_{x_t}^{h,w}$ is not well defined in Eq.~\eqref{eq:y}.
Thus, we introduce our binary selection mask $T$, defined as:
\begin{equation}
    T^{h,w} = \mathbbm{1}\left(\sum_c \hat{p}_{x_t}^{h,w,c} > 0\right),    \label{eq:T}
\end{equation}
so that $T^{h,w}\cdot y_{x_t}^{h,w,c} $ (i.e., the final reliable pseudo label $R_{x_t}^{h,w,c}$) is well defined. }
For voxel $(h,w)$, if the revised probability vector $\hat{p}_{x_t}^{h,w}$ is a zero vector, then $T^{h,w}=0$, 
and the voxel $(h,w)$ is not selected in the segmentation loss in Eq.~\eqref{eq:loss}. 

It should be noted that the double-threshold selection scheme could alleviate the class imbalance between hard and easy-to-transfer classes. 
For example, an abdominal CT image contains $80\%$ backgrounds (treated as class $0$), $10\%$ spleen area  (treated as class $1$) and $10\%$ liver area (treated as class $2$). 
For a voxel $(h,w)$, its pseudo label probability $p_{x_t}^{h,w,c}$ may be predicted as the background class due to class imbalance.
Then, after the double-threshold selection, the revised value $\hat{p}_{x_t}^{h,w,0}$ may be set to $0$, 
so that the pseudo target label $y_{x_t}^{h,w,c}$ may point to the other two minority classes. 
Reliable pseudo label selection using the binary mask $T$ and the revised pseudo target label $y_{x_t}$ overcomes the class imbalance issue in SFUDA. 


\subsubsection{Prototype Based Selection}

The above double-threshold selection performs pseudo label selection by considering global probability information and intra-class information, 
while it does not consider inter-class information. 
Here we propose a prototype based pseudo label selection method by considering inter-class information. 


For a target image $x_t \in \mathbb{R}^{H\times W \times C}$, its feature map after the backbone of the model (e.g., DeepLabV3~\cite{chen2017deeplab}) is resized to the size $H\times W \times L$ by bilinear interpolation, 
where $L$ is the channel of the feature map. 
Denote the resized feature map as $e\in \mathbb{R}^{H\times W \times L}$, and the feature vector for voxel $(h,w)$ is $e^{h,w} \in \mathbb{R}^L$. 
Then, we define the prototype for the $c$-th class as
\begin{equation}
  z^{c}= \frac{\sum_{h,w}(e^{h,w} \cdot \hat{p}_{x_t}^{h,w,c})}{\sum_{h,w} \hat{p}_{x_t}^{h,w,c}}, 
\end{equation}
which is a weighted mean of the feature vectors of all voxels, and the weights are the revised pseudo label $\{p_{x_t}^{h,w,c}\}$ in Eq.~\eqref{eq:threshold}.
In this way, the prototype gives more weighting for reliable voxels with reliable probabilities.

Then, for voxel $(h,w)$, the feature distance between the feature vector $e^{h,w}$ and the prototype of the $c$-th class is:
\begin{equation}
  d^{h,w,c}= ||e^{h,w}-z^c||_2.
\end{equation}
Denote the $N_c$ dimensional vector of distances for all $N_c$ classes as $d^{h,w}=[d^{h,w,0},\cdots, d^{h,w,N_c-1} ]^T$. 
For voxel $(h,w)$, we could predict its class label based on the nearest feature distance among $d^{h,w}$, i,e., $\argmax(d^{h,w})$. 
If this prediction is consistent with the one-hot pseudo target label $y_{x_t}^{h,w}$, 
then the pseudo label $y_{x_t}^{h,w}$ is likely reliable. 
Otherwise, if these two predictions are not consistent, 
then we do not consider the voxel in the segmentation loss in Eq.~\eqref{eq:loss}. 
Therefore, we could define the binary selection mask $T^{h,w}$ as:
\begin{footnotesize}

\begin{equation}
      T^{h,w}= \mathbbm{1}\left(\sum_c \hat{p}_{x_t}^{h,w,c} > 0\right) \mathbbm{1}\left(\argmax(y_{x_t}^{h,w}) =={\argmin} (d^{h,w})\right). \label{eq:T_2}
\end{equation}
\end{footnotesize}

With the proposed binary mask $T$ in Eq.~\eqref{eq:T_2} or Eq.~\eqref{eq:T}, and the pseudo target label $y_{x_t}$ in Eq.~\eqref{eq:y}, 
the segmentation loss in Eq.~\eqref{eq:loss} could train the FVP $v_f$ using the unlabeled target data and the frozen pre-trained model.

\section{Experiments and results}

\begin{table*}
  \caption{\label{tab:Abdominal}\textbf{Quantitative evaluation results of the segmentation on the Abdominal dataset.} 
  The $\uparrow$ sign indicates a higher score is better, while the $\downarrow$ sign indicates a lower score is better.  
  The best results are in boldface.}
\centering
\setlength{\tabcolsep}{3pt}
\begin{tabular}{cccccc|cccccc|}
\toprule
\multicolumn{1}{c}{\multirow{2}{*}{\textbf{Method (CT$\rightarrow$MRI)}}}

& \multicolumn{5}{c|}{\textbf{Dice} $\uparrow$} 

& \multicolumn{5}{c}{\textbf{ASD} $\downarrow$} 
\\
\cmidrule(l){2-11}

\multicolumn{1}{c}{}
                   
& \textbf{Liver} & \textbf{R.kidney}  & \textbf{L.kidney} & \textbf{Spleen} &\textbf{Average} & \textbf{Liver} & \textbf{R.kidney}  & \textbf{L.kidney} & \textbf{Spleen} &\textbf{Average}\\  

\hline
source-only & 0.582 & 0.789 & 0.686 & 0.009 & 0.517
& 4.661 & 2.678 & 1.545 & 14.518 & 5.850
 \\
target supervised&0.727& 0.941&  0.941& 0.713&0.831
& 3.250& 0.4840& 0.272& 6.120&2.532\\
\hline
LD\cite{you2021domain}& 0.627 & 0.867 & 0.784 & 0.482 & 0.690
  & \textbf{4.198} & 2.163 & 1.718 & 6.785 & 3.716
 \\
DPL\cite{chen2021source} & 0.556 & 0.860 & 0.785 & 0.476 & 0.669
  &4.429 & \textbf{2.048} & 1.623 & 7.576 & 4.669
 \\
OS\cite{liu2021adapting} &0.556 & 0.854 & 0.797 & 0.532 & 0.685
&4.516 & 2.225 & 1.676 & 6.626 & 3.761
 \\
SFDA\cite{SFDA}&0.521&0.812&0.712&0.448&0.623&5.188&2.589&\textbf{1.414}&8.779&4.493\\
FSM\cite{FSM} &0.632&0.854&0.796&0.508&0.698&4.770&2.546&1.721&6.755&3.948\\
  FVP (ours) &\textbf{0.648}&\textbf{0.876} & \textbf{0.803} & \textbf{0.605} &\textbf{0.733}& 4.483& 2.101&1.542 &\textbf{6.153}&\textbf{3.570} \\
 \bottomrule
\end{tabular}

\begin{tabular}{cccccc|cccccc|}
\multicolumn{1}{c}{\multirow{2}{*}{\textbf{Method (MRI$\rightarrow$CT)}}}

& \multicolumn{5}{c|}{\textbf{Dice} $\uparrow$ } 

& \multicolumn{5}{c}{\textbf{ASD} $\downarrow$} 
\\
\cmidrule(l){2-11}

\multicolumn{1}{c}{}
                   
& \textbf{Liver} & \textbf{R.kidney}  & \textbf{L.kidney} & \textbf{Spleen} &\textbf{Average} & \textbf{Liver} & \textbf{R.kidney}  & \textbf{L.kidney} & \textbf{Spleen} &\textbf{Average}\\  

\hline

source-only & 0.842 & 0.607 & 0.622 & 0.520 & 0.647&
6.628 & 5.141 & 5.067 & 6.914 & 5.938
 \\

 target supervised &0.961& 0.917&  0.915& 0.945&0.934 
&1.071& 1.052&1.363& 0.644&1.033\\
\hline
  LD\cite{you2021domain}& 0.758 & \textbf{0.680} &\textbf{0.751} & 0.580 & 0.692
  &6.422 & 4.696 & \textbf{1.992} & 5.627 & 4.684
 \\
DPL\cite{chen2021source} &0.844 & 0.603 & 0.569 & 0.598 & 0.653
  &4.171 & \textbf{2.521} & 2.200 & 3.486 & 3.095
 \\
OS\cite{liu2021adapting} &0.835 & 0.607 & 0.570 & 0.620 & 0.658
&4.677 & 3.228 & 2.242 & 3.809 & 3.489
 \\
 SFDA\cite{SFDA}&0.703&0.559&0.589&0.511&0.591&5.128&5.287&4.331&6.785&5.383\\
 FSM\cite{FSM} &0.870&0.619&0.694&0.688&0.718&4.584&4.696&3.902&4.113&4.324\\
  FVP (ours) & \textbf{0.878}&0.647 & 0.732&\textbf{0.683}&\textbf{0.735}& \textbf{3.631}& 2.583&3.102 &\textbf{2.336} & \textbf{2.913} \\

 \bottomrule

\end{tabular}

\end{table*}

\begin{table*}
  \caption{\label{tab:MMWHS}\textbf{Quantitative evaluation results of the segmentation on the MM-WHS dataset.}  
  The best results are in boldface.}
\centering
\setlength{\tabcolsep}{3pt}
\begin{tabular}{cccccc|cccccc|}
\toprule
\multicolumn{1}{c}{\multirow{2}{*}{\textbf{Method (CT$\rightarrow$MRI)}}}

& \multicolumn{5}{c|}{\textbf{Dice} $\uparrow$} 

& \multicolumn{5}{c}{\textbf{ASD} $\downarrow$} 
\\
\cmidrule(l){2-11}

\multicolumn{1}{c}{}

& \textbf{AA} & \textbf{LAC}  &\textbf{ LVC} & \textbf{MYO} &\textbf{Average} & \textbf{AA} & \textbf{LAC}  & \textbf{LVC} & \textbf{MYO} &\textbf{Average}\\ 

\hline
source-only &0.390 & 0.281 & 0.514 & 0.464 & 0.412&13.679& 32.500 & 25.591 & 18.232 & 22.501

\\

target supervised&0.800 &0.871 & 0.930& 0.874 & 0.869&1.374& 1.312 & 1.013 & 2.144 & 1.461
\\
\hline
LD\cite{you2021domain}& 0.573 & 0.247 & 0.492 & 0.473 & 0.446&\textbf{10.503 }& 28.620 & 26.740 & 18.287 & 21.038
 \\
DPL\cite{chen2021source} &0.574 & 0.249& 0.514 &\textbf{0.508} & 0.461&12.478 & 28.551 & \textbf{18.827 }& 15.522 & 20.094\\
OS\cite{liu2021adapting}  &\textbf{0.576} & 0.247 & 0.532 & 0.513 & 0.467&11.671 & 28.150 & 24.097 & 15.097 & 19.754\\ 
SFDA\cite{SFDA}&0.422&0.289&0.568&0.466&0.436&12.421&31.298&24.420&15.752&20.973\\
FSM\cite{FSM} &0.504&0.413&0.517&0.449&0.471&12.460&27.092&23.758&17.883&20.300\\
FVP (ours) &0.385&\textbf{0.448}&\textbf{0.578}&  0.491&
\textbf{0.476} &19.012 &\textbf{24.661 }&18.923& \textbf{14.559}&\textbf{19.289} \\
 \bottomrule
\end{tabular}

\begin{tabular}{cccccc|cccccc|}
\multicolumn{1}{c}{\multirow{2}{*}{\textbf{Method (MRI$\rightarrow$CT)}}}

& \multicolumn{5}{c|}{\textbf{Dice} $\uparrow$} 

& \multicolumn{5}{c}{\textbf{ASD} $\downarrow$} 
\\
\cmidrule(l){2-11}

\multicolumn{1}{c}{}
                   
& \textbf{AA} & \textbf{LAC}  & \textbf{LVC} & \textbf{MYO} &\textbf{Average} & \textbf{AA} & \textbf{LAC}  & \textbf{LVC} & \textbf{MYO} &\textbf{Average}\\ 

\hline

source-only &0.861 & 0.576 & 0.774 & 0.646 & 0.714&12.729 & 11.746 & 6.356 & 4.866 & 8.924\\
target supervised &0.892 &0.916 & 0.933& 0.948 &0.922&1.220 & 2.169& 1.071& 0.781&1.310\\
\hline
LD\cite{you2021domain}& 0.878 & \textbf{0.734} &\textbf{0.814} & \textbf{0.725 }&\textbf{0.788}&12.754 & 9.450 &\textbf{3.758} & \textbf{3.490 }& 7.363\\
DPL\cite{chen2021source} &\textbf{0.910 }& 0.694 & 0.782 & 0.652 & 0.760 & \textbf{8.555}&\textbf{9.000}& 6.109 & 4.853 &7.129\\
OS\cite{liu2021adapting}& 0.891 & 0.677 & 0.766 & 0.659 & 0.748 &  9.021 & 9.412 & 9.918 & 6.652 & 8.750\\
SFDA\cite{SFDA}&0.855&0.628&0.757&0.656&0.724&13.228&9.986&7.183&4.537&8.734\\
FSM\cite{FSM} &0.849&0.616&0.779&0.673&0.729&10.394&10.165&7.774&5.329&8.416\\
FVP (ours) & 0.856& 0.719 & 0.795&0.640&0.753& 9.012& 9.003&4.374 &3.520 &\textbf{6.477} \\

 \bottomrule
\end{tabular}
\end{table*}

\subsection{Data and Experimental Setup}

We validate our proposed FVP method for SFUDA of medical image segmentation by using three public datasets.

\subsubsection{the Abdominal Dataset}
This dataset is widely used in domain adaptation tasks~\cite{SIFA,hoffman2018cycada,tomar2021self}, 
which contains two groups of the Abdominal data: 
20 MRI scans from the CHAOS challenge~\cite{kavur2021chaos} and 30 CT scans from Multi-Atlas Labeling Beyond the Cranial Vaulti-Workshop and Challenge~\cite{landman2015multi}. 
The dataset has labels of four organs: liver, right kidney (R-Kid), left kidney (L-Kid) and spleen. 
The size of each MRI scan is $256 \times 256 \times L$ ($21 \leq L \leq 30$) within a 3D volume, 
where $L$ is the length of the long axis and is different in subjects. 
Each CT scan is $512 \times 512 \times L$ and we crop the image into $256 \times 256 \times L$ ($35 \leq L \leq 117$). 
We randomly divide the training set and test set with the ratio of $4:1$ for each modality.

\subsubsection{the MM-WHS Dataset} 
There are unpaired $20$ MRI and $20$ CT 3D images with golden standard segmentation labels in the \textit{Multi-Modality Whole Heart Segmentation Challenge 2017} dataset \cite{MMWHS}. 
All CT data cover the whole heart from the upper abdominal to the aortic arch and the slices were acquired in the axial view.  The inplane resolution is about $0.78 \times 0.78$ mm and the average slice thickness is $1.60$ mm. The MRI data were acquired using 3D balanced steady-state free precession (b-SSFP) sequences with about $2$ mm acquisition resolution at each direction and were resampled into about $1$ mm.
The ground truth labels consist of the ascending aorta (AA), the left atrium blood cavity (LAC), the left ventricle blood cavity (LVC), 
and the myocardium of the left ventricle (MYO). 
Each modality is randomly split as training and test sets with the ratio of $4:1$. 
We manually cropped the original scans with a fixed coronal plane size of $256 \times 256$, as the same way in~\cite{SIFA}.

\subsubsection{the BraTs 2018 Dataset}
{ Multi-modality brain tumor segmentation challenge 2018 dataset (BraTS 2018) \cite{6975210} is a dataset that provides multimodal 3D brain MRIs and ground truth segmentations, consisting of 4 MRI modalities per case (T1, T1c, T2 and FLAIR). The dataset contains 75 patient data. In the original dataset of BraTS, the segmentation task involves three distinct annotations: GD-enhancing tumor (ET), peritumoral edema (ED), and necrotic and non-enhancing tumor core (NCR/NET). These annotations are labeled as 4, 2, and 1, respectively, while everything else is labeled as 0, following the guidelines outlined in the BraTS reference paper\cite{6975210}. When applying our method to the BraTS2018 dataset, we simplify the segmentation task to focus solely on the whole tumor (WT), which encompasses labels 1, 2, and 4, along with the background labeled as 0. This setting is consistent with common segmentation settings found in Unsupervised Domain Adaptation (UDA) articles\cite{xie2022unsupervised,zhou2022generalizable}. We demonstrate our method between two modalities of MRI imaging from low-graded glioma cases: FLAIR and T2. We randomly split the training set and the test set by 4:1. Additionally, each axial slice is resized to $192 \times 168$.
 }

\subsubsection{Evaluation Metrics}
One of the most commonly-used evaluation metrics to quantitatively evaluate the performance of the segmentation is the Dice coefficient. 
It calculates the overlap region between the prediction masks and the ground truths normalized by the sum of the prediction and ground truth regions. 
In our experiments, we calculate the Dice values in 3D volumes, and the higher Dice indicates a better performance. 
Another metric we employ is Average Surface Distance (ASD) for boundary agreement assessment, which is the lower the better.

\subsubsection{Implementation Details}

We adopt the DeepLabV3~\cite{chen2017deeplab} with resnet50 backbone~\cite{he2016deep} as our segmentation model for training the pre-trained source model $S$. 
All the data is pre-processed into 0-mean and 1-variance. 
Given a target images $x_t$, we first directly inference predictions by $S$ as the initial pseudo label $p_{x_t}$. 
Then, we adopt the reliable label detection module to select reliable pseudo labels using a binary mask $T$ and revised pseudo target label $y_{x_t}$, as shown in Section~\ref{sec:ST}. 
In our comparison experiments, 
we use $k=0.8$ to set intra-class threshold $\delta_{x_t}^c$ as the top $80\%$ of the probability values in $p_{x_t}^{c}$, 
and set the global probability threshold $\lambda=0.01$. These two values are set empirically in our experiments. We do not specifically optimize these parameters for different datasets and simply set the same values for all datasets.
The Fourier visual prompt $v_f$ is trained by using the segmentation loss in Eq.~\eqref{eq:loss} with these reliable pseudo labels, 
the optimizer is set as Adam with the weight decay as $1e^{-5}$ and batch size as $8$. 
The learning rates are tuned with $\{0.01, 0.1, 1\}$ for different experiments.  
All the experiments are performed with Pytorch 1.10.1 using NVIDIA GPU V100.

\subsection{Comparison Experiments}
\label{sota}

\subsubsection{Experiment Setting}

We compare our proposed FVP with five state-of-art methods in SFUDA, 
including Label-denoising framework (LD)~\cite{you2021domain}, 
Denoised Pseudo-Labeling method (DPL)~\cite{chen2021source}, 
"off-the-shelf" segmentation model (OS)~\cite{liu2021adapting}, 
SFDA~\cite{SFDA}, 
and Fourier Style Mining (FSM)~\cite{FSM}.
LD~\cite{you2021domain} integrates positive and negative learning together with pseudo labeling. 
DPL~\cite{chen2021source} adopts denoising pseudo labeling via prototypes and uncertainty estimation. 
OS~\cite{liu2021adapting} uses the adaptation of the batch normalization layer to achieve domain-wise alignment. 
SFDA~\cite{SFDA} enables to recover and preserve the source domain knowledge from the pre-trained source model and distills target domain information for self-supervised training. 
FSM~\cite{FSM} is one of the latest published SOTA method, 
which is composed of two stages: generation and adaptation. 
FSM firstly generates source-like images through statistic information of the pre-trained source model and mutual Fourier Transform, 
and then achieve feature-level adaptation via a Contrastive Domain Distillation~\cite{FSM}.
It should be noted that all these SFUDA methods need to modify the pre-trained model or train a new model, 
while our proposed FVP perform SFUDA by using a visual prompt, without requiring changes of the pre-trained model, as shown in Fig.~\ref{fig:SFUDA}(b) and (c). 

We implement these methods based on their papers and their released codes on the Abdominal and the MM-WHS datasets for a fair comparison. 
"Source-only" means directly applying the source model to the target data without any adaptation, 
which is shown as a baseline. 
We perform the SFUDA task with two adaptation directions using these two modalities: 
one is CT$\rightarrow$MRI which means from CT (as the source domain) to MRI (as the target domain), 
and the other one is MRI$\rightarrow$CT which means from MRI (as the source domain) to CT (as the target domain). For the BraTS dataset, the two adaptation directions are T2$\rightarrow$Flair and Flair$\rightarrow$T2. 

\subsubsection{the Abdominal Results}

\begin{figure*}[t]
     \raggedleft
    \includegraphics[width=1\linewidth]{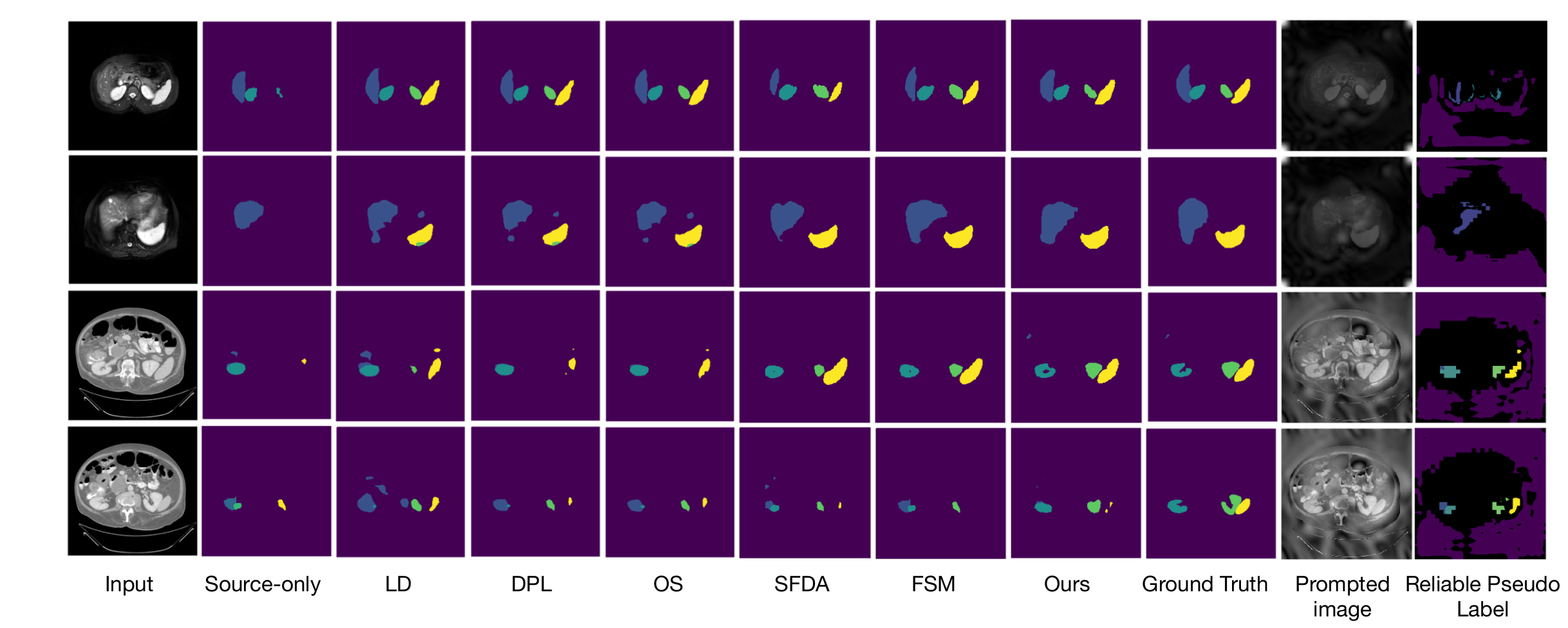}
    \caption{\label{fig:visual_abdominal_seg}\textbf{Visualization of the segmentation results on the Abdominal dataset.} 
    The structure of the Liver, R.kidney, L.kidney, and Spleen are shown in blue, dark green, light green, and yellow colors, respectively. 
    The first two rows are segmentation results of MR images in CT$\rightarrow$MRI. 
    The last two rows are segmentation results of CT images in MRI$\rightarrow$CT. 
    In each row, we visualize the input image, segmentation results (source-only, LD, DPL, OS, SFDA, FSM, our FVP), ground truths, prompted images, and the reliable pseudo labels, respectively. 
    Unreliable voxels are shown in dark in the reliable pseudo label images.
    }
\end{figure*}

\begin{figure*}[t]
    \includegraphics[width=1\linewidth]{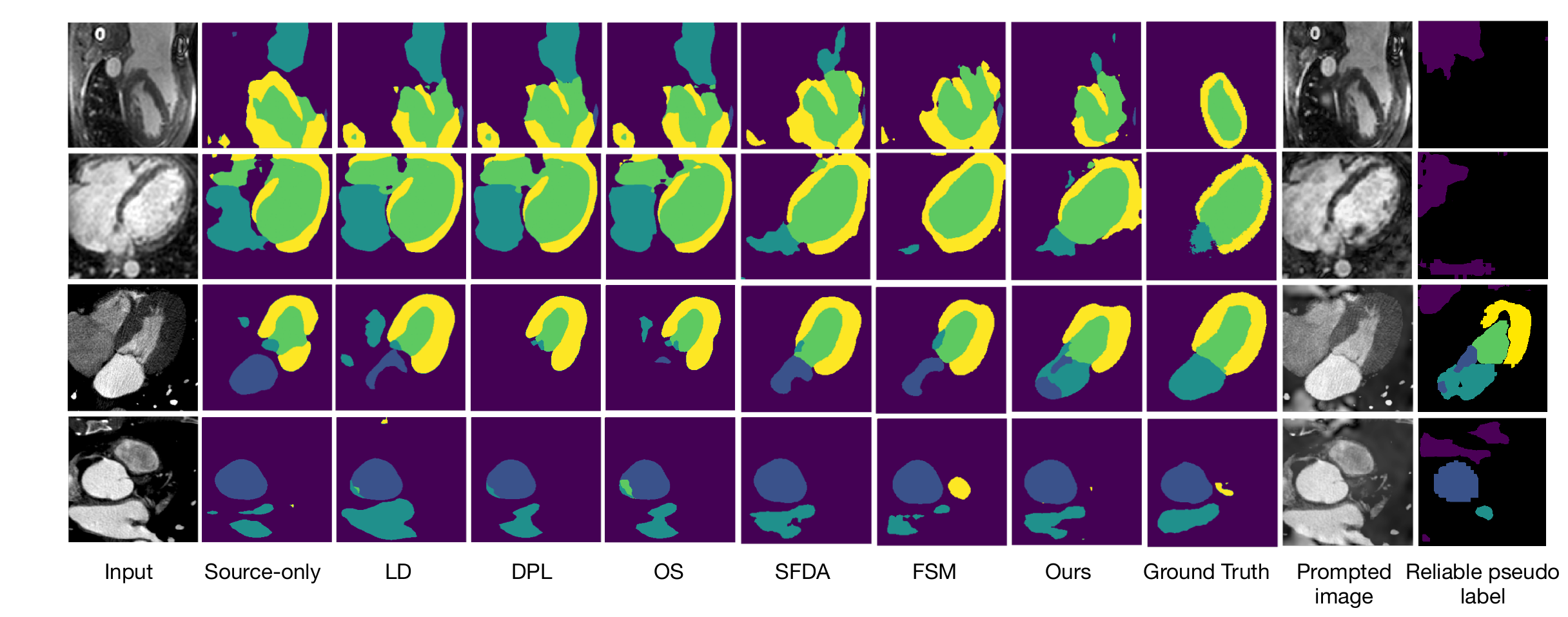}
    \caption{\label{fig:visual_cardiac_seg}\textbf{Visualization of the segmentation results on the MM-WHS dataset.} 
    The structure of the LA, AA, LV and MYO are shown in blue, dark green, light green, and yellow colors, respectively. 
    The first two rows are segmentation results of MR images in CT$\rightarrow$MRI. 
    The last two rows are segmentation results of CT images in MRI$\rightarrow$CT. 
    }
\end{figure*}
\begin{figure*}[t]
    \includegraphics[width=1\linewidth]{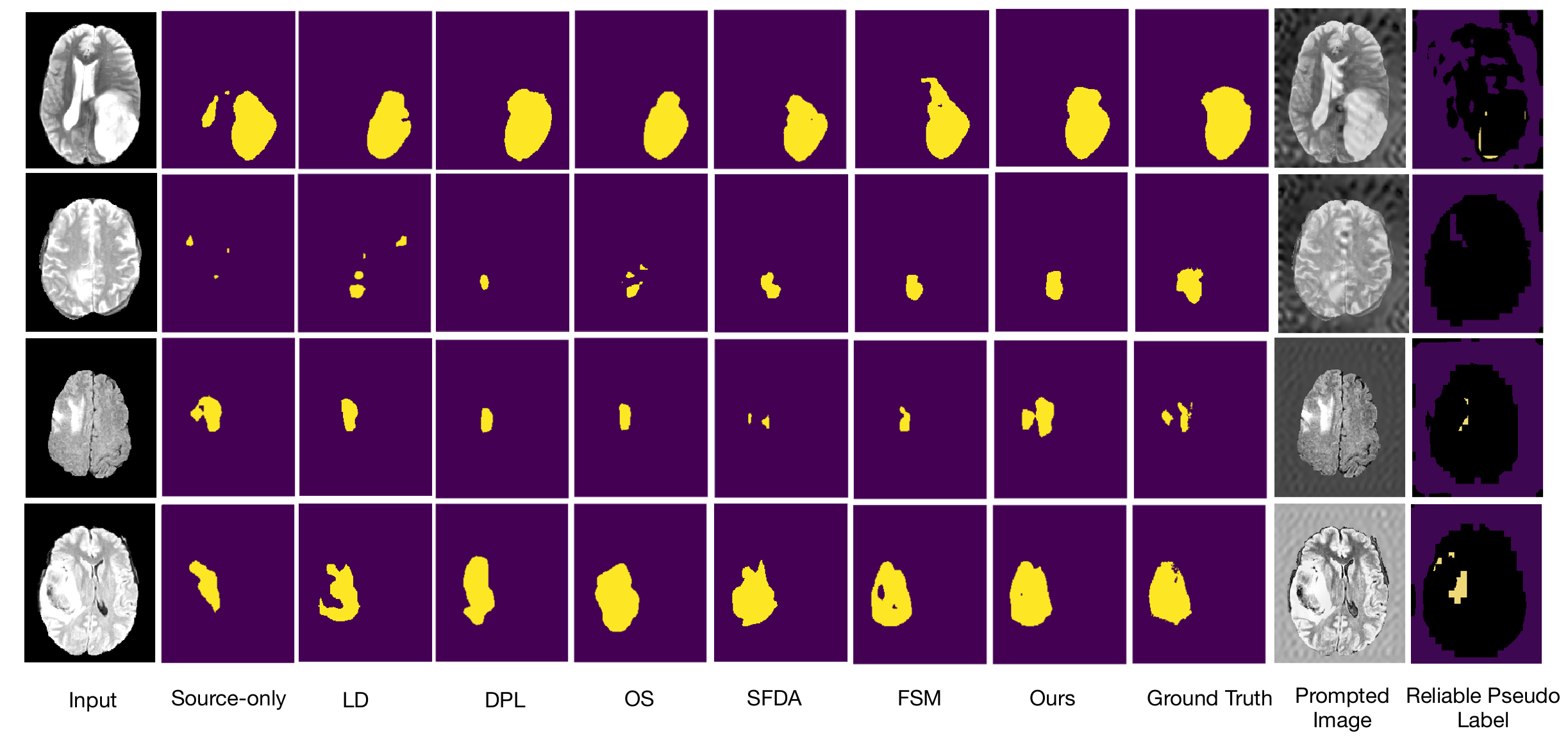}
    \caption{\label{fig:visual_b rats_seg}\textbf{Visualization of the BraTS 2018 dataset segmentation results.} 
    The first two rows are segmentation results of T2 images in Flair$\rightarrow$T2. 
    The last two rows are segmentation results of Flair images in T2$\rightarrow$Flair. 
    }
\end{figure*}

The quantitative comparison results of the Abdominal dataset are shown in Table~\ref{tab:Abdominal}. 
For this dataset, the number of CT samples (30) is larger than the number of MRI samples (20). 
While the Dice score by source-only in CT$\rightarrow$MRI ($0.517$) is much lower than that in MRI$\rightarrow$CT ($0.647$), 
which indicates that the adaptation of CT$\rightarrow$MRI is more difficult than the opposition MRI$\rightarrow$CT. 
It is probably because that MRI provides more texture details in organs, compared with CT. 
In both adaptation directions, our proposed FVP yields generally the best performance with the highest average Dice ($0.733$, $0.735$) and lowest average ASD ($3.570$, $2.913$). 
Especially for the hard adaptation direction CT$\rightarrow$MRI, FVP obtains best Dice in all four organs. 
The spleen is the hardest adaptation organ in CT$\rightarrow$MRI, as source-only obtains a Dice of $0.009$. 
While our proposed FVP yields the best Dice $0.605$ for the spleen in CT$\rightarrow$MRI. 
The visualization of some segmentation results are shown in Fig.~\ref{fig:visual_abdominal_seg}.

It should be noted that our FVP achieves the goal of SFUDA with only a very small number of trainable parameters, 
because the pre-trained model is frozen. 
To be specific, the number of trainable parameters is proportional to the size of the visual prompt during training. 
While the numbers of trainable parameters in the other methods are generally proportional to the size of the pre-trained source model, 
because the pre-trained model is trained in the adaptation. 
Moreover, some methods introduced additional networks for adaptation. 
The comparison of the model parameters is shown in Table~\ref{tab:number}, 
where our prompt size is designed to $32 \times 32$ and we apply the same segmentation model (DeepLabV3) for a fair comparison. 
We also compare the training time for one epoch on one NVIDIA GPU V100, our method speeds up about 10 times compared to the other SFUDA methods. 
For the FSM method, we only compare its second adaptation stage.

 \begin{table}
  \caption{\label{tab:number}\textbf{Comparison of the number of the trainable parameters and training time}.}
\centering
\setlength{\tabcolsep}{3pt}
\begin{tabular}{c|c|c}
\toprule
Method& Number of parameters &Training time for 1 epoch  \\

\hline source-only &-&-\\   
LD\cite{you2021domain}& $3.96 \times 10^7$ &17.56s
 \\
DPL\cite{chen2021source}& $3.96 \times 10^7$ &20.51s
 \\
OS\cite{liu2021adapting}& $3.96 \times 10^7$ &30.12s\\
SFDA\cite{SFDA}& $4.76 \times 10^7$ &20.18s\\
FSM\cite{FSM}& $7.92 \times 10^7$ &25.51s\\
FVP (ours)&2048&2.93s\\
 \bottomrule
\end{tabular}

\end{table}

\subsubsection{the MM-WHS Results}

Table~\ref{tab:MMWHS} shows the quantitative results of different methods for the MM-WHS dataset. 
Similarly with Table~\ref{tab:Abdominal}, the Dice score by source-only in CT$\rightarrow$MRI ($0.412$) is much lower than that in MRI$\rightarrow$CT ($0.714$),  
which also indicates that the adaptation of CT$\rightarrow$MRI is more difficult than MRI$\rightarrow$CT. 
For CT$\rightarrow$MRI, our FVP yields the best performance with the highest average Dice ($0.476$), and the lowest average ASD ($19.289$). 
For MRI$\rightarrow$CT, our FVP provides comparable average Dice ($0.753$) with the highest Dice ($0.788$) by LD, higher than the Dice of source-only ($0.714$), 
and FVP yields the best average ASD ($6.477$), much lower than ASD by LD ($7.363$), and ASD by source-only ($8.924$). 
The visual comparison of cardiac segmentation results is shown in Fig.~\ref{fig:visual_cardiac_seg}.

\subsubsection{the BraTS 2018 Results}
{Table~\ref{BraTS} shows the results of the proposed FVP and comparative methods for the BraTS 2018 dataset. 
The proposed FVP improves the Dice performance from $0.656$ to $0.697$ and $0.770$ to $0.813$ in two adaptation tasks respectively, which surpasses five comparative methods. 
Fig.~\ref{fig:visual_b rats_seg} shows some visualization results, 
which demonstrates that the segmentation results by our method are closer to the ground truth, compared with other methods.
}


\begin{table}
  \caption{\label{tab:Multi-level-bra}\textbf{Quantitative evaluation results of the segmentation on the BraTS 2018 dataset}.}
\centering
\setlength{\tabcolsep}{3pt}
\begin{tabular}{c|c|c|c|c}
\toprule
\multirow{2}{*}{\makecell{}{}{Method}}&\multicolumn{2}{c|}{\makecell{FLAIR $\rightarrow$ T2}{}{}}& \multicolumn{2}{c}{\makecell{T2 $\rightarrow$ FLAIR}{}{}} \\
\cline{2-5}
& Dice&ASD&Dice&ASD\\
\hline source-only &0.656 &5.520  &0.770  &3.718\\  
target supervised &0.859 &2.393  &0.886  &1.483\\ 
\hline
LD\cite{you2021domain}& 0.667   & 5.502    &   0.795   &3.044
 \\
DPL\cite{chen2021source}&0.685    &   4.743  &   0.806   & 3.213
 \\
OS\cite{liu2021adapting}&  0.679    & 4.694   & 0.799&3.028\\
SFDA\cite{SFDA}&0.674 &5.118 &0.792&2.970 \\
FSM\cite{FSM} &0.679&5.236&0.810&2.811\\
FVP (ours)& \textbf{0.697}    & \textbf{4.472}     &\textbf{0.813}   & \textbf{2.709}\\
 \bottomrule
\end{tabular}
\label{BraTS}
\end{table}

\subsection{Ablation Study}
We conduct ablation experiments to investigate the design of Fourier visual prompt and the effect of the reliable label detection module.

\subsubsection{Design of Fourier Visual Prompt}


Existing works in~\cite{FDA,yang2020phase,fourierDG} have demonstrated that 
the low frequency part of images plays an important role in style transfer and domain adaptation, 
and they proposed to modify only the amplitude components of images for adaptation. 
In this paper, we propose to parameterize the visual prompt using its low-frequency components,  
so that it can change the input image style in a global manner with less learnable parameters.  
We propose both the complex FVP in~\eqref{eq:fvp_complex} to change both amplitude and phase components 
and the real FVP in~\eqref{eq:fvp_amplitude} to change only the amplitude. 
In this subsection, we would like to investigate whether changing the phase component is still useful for domain adaptation. 

\begin{figure}[t]
    \centering
    \includegraphics[width=1.0\linewidth]{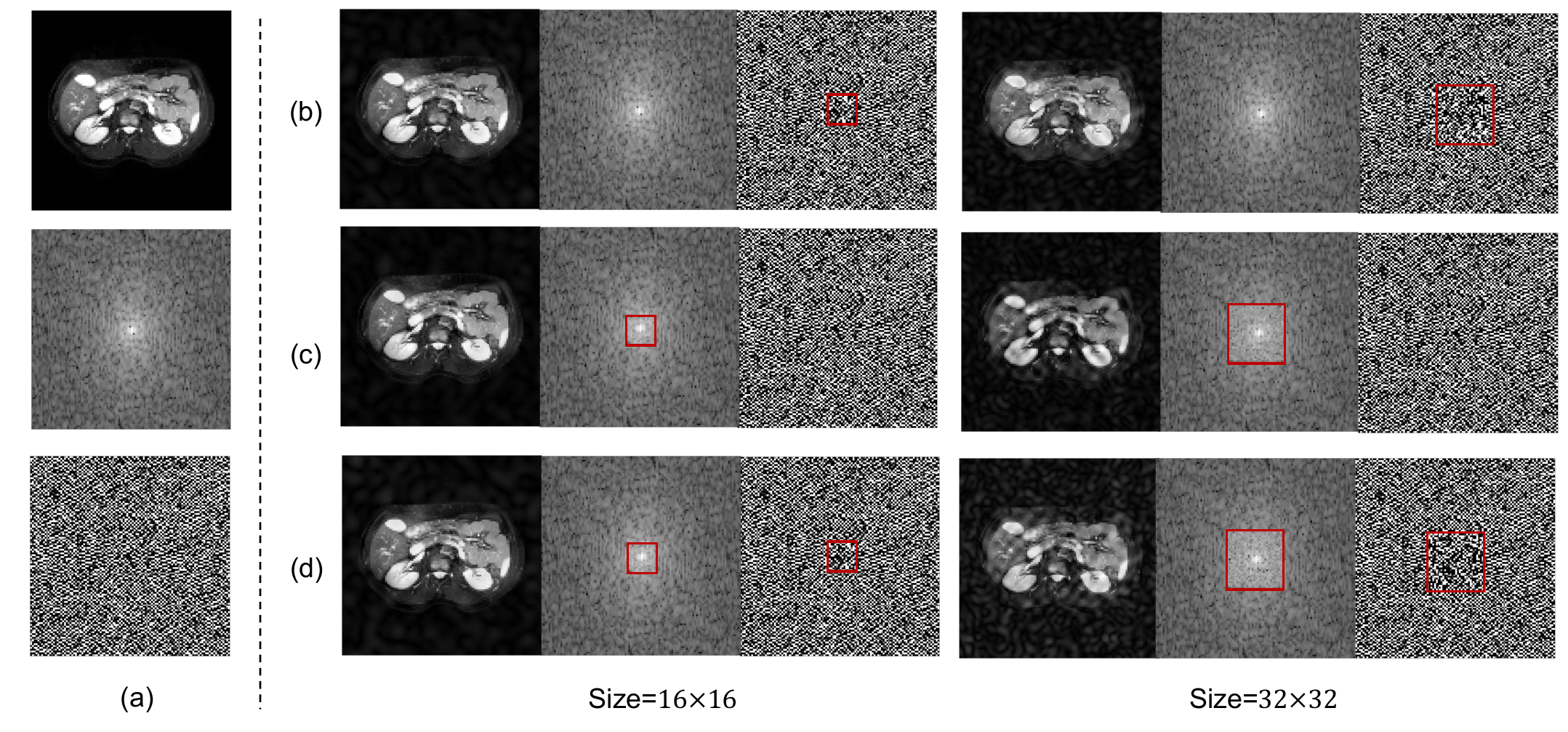}
    \caption{\label{fig:change_amp_pha}\textbf{Change amplitude and phase components of an image with different prompt sizes.} 
    (a) An image with its amplitude and phase components. 
    (b) The image reconstructed by the noisy phase component and the original amplitude. 
    (c) The image reconstructed by the noisy amplitude component and the original phase. 
    (d) The image by adding a zero mean complex Gaussian noise to the low-frequency part of the image.}
\end{figure}

Fig.~\ref{fig:change_amp_pha} (a) shows an image with its phase and amplitude components. 
Then, in Fig.~\ref{fig:change_amp_pha} (d) we add a zero mean complex Gaussian noise to the low-frequency part (with size $16\times 16$ or $32\times 32$) of the image, 
which changes both phase and amplitude components simultaneously. 
In this way, the Fourier visual prompt is simulated as a complex noise image. 
Fig.~\ref{fig:change_amp_pha} (b) shows the image reconstructed by the noisy phase component and the original amplitude. 
Fig.~\ref{fig:change_amp_pha} (c) shows the image reconstructed by the noisy amplitude component and the original phase. 
All these images demonstrate meaningful semantic information of boundaries of organs, although they have different image appearances due to the noise-simulated FVPs.
It indicates that changing the phase component of images could also be helpful for domain adaptation. 
Moreover, the complex FVP in Eq.~\eqref{eq:fvp_complex} is much more computationally efficient than the real FVP in Eq.~\eqref{eq:fvp_amplitude}, 
because it requires much less number of FFT computation as we analyzed in Section~\ref{sec:FVP}.
Last, when using the complex FVP, we could let the loss in Eq.~\eqref{eq:loss} and the training process automatically learn both the amplitude and phase components of FVP, 
instead of manually setting the phase component of FVP to zero. 


Besides the experiments of simulated FVPs in Fig.~\ref{fig:change_amp_pha}, 
Table \ref{RealFVP} shows the domain adaptation results of the Abdominal dataset by using complex FVP and two real FVPs which change the amplitude and phase components respectively. 
As shown in Table \ref{RealFVP}, 
the real FVP in the amplitude component has higher average Dice values than the real FVP in the phase component, 
which means the amplitude component of the input image is more suitable for adaptation. 
It is consistent with existing works in~\cite{FDA,yang2020phase,fourierDG}.  
However, compared with two real FVPs, the complex FVP yields the best adaptation performance with the highest average Dice values in both adaptation directions (CT$\rightarrow$MRI and MRI$\rightarrow$CT).

\begin{table}
{
  \caption{\textbf{Domain adaptation results of the Abdominal dataset by using the complex and real FVPs.} 
  "Amplitude" means the real FVP changes only the amplitude component of the input image. 
  "Phase" means the real FVP to change only the amplitude component. 
  "Complex" means the complex FVP changes both the amplitude and real components.}
\label{RealFVP}}
\centering
\setlength{\tabcolsep}{3pt}
\begin{tabular}{cccccc}
\toprule
Method (CT$\rightarrow$MRI) & \textbf{Liver} & \textbf{R.kidney}  & \textbf{L.kidney} & \textbf{Spleen} &\textbf{Average} \\  

\hline
Amplitude  & \textbf{0.701} & 0.840 & 0.746 & 0.136 &0.606

 \\
Phase& 0.608 & 0.742&  0.658& 0.245&0.563

 \\
Complex & 0.648 & \textbf{0.876} & \textbf{0.803} & \textbf{0.605} & \textbf{0.733}

\\
 \bottomrule
\end{tabular}
\begin{tabular}{cccccc}

Method (MRI$\rightarrow$CT) & \textbf{Liver} & \textbf{R.kidney}  & \textbf{L.kidney} & \textbf{Spleen} &\textbf{Average} \\  

\hline
Amplitude  &0.852 & \textbf{0.703} & 0.671 & 0.636 &0.716

 \\
Phase& 0.850 & 0.645&  0.669& 0.573&0.684

 \\
Complex &  \textbf{0.878}&0.647 & \textbf{0.732}&\textbf{0.683}&\textbf{0.735}

\\
 \bottomrule
\end{tabular}

\end{table}

\begin{figure}[t]
    \centering
    \subfigure[CT$\rightarrow$MRI]{
    \begin{minipage}[t]{0.48\linewidth}
    	\begin{center}
    		\includegraphics[width=1\linewidth]{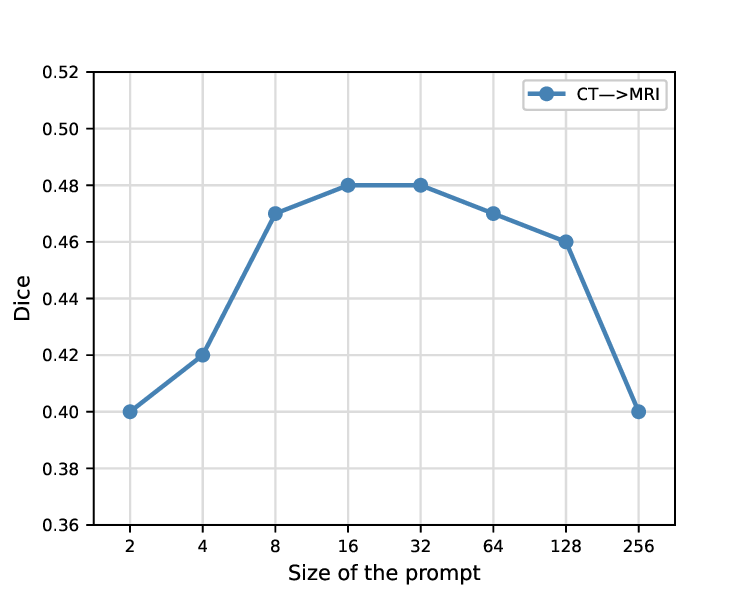}
    	\end{center}
    \end{minipage}%
    }%
    \subfigure[MRI$\rightarrow$CT]{
    \begin{minipage}[t]{0.48\linewidth}
    	\begin{center}
    		\includegraphics[width=1\linewidth]{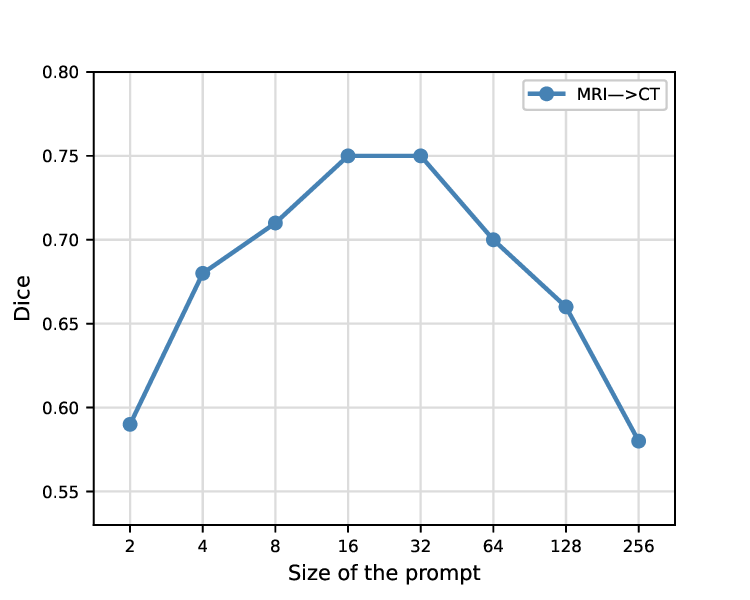}
    	\end{center}
    \end{minipage}%
    }%
    \caption{\label{fig:ablation_size}\textbf{Effect of different prompt sizes}. 
    Dice scores on the MM-WHS dataset with different size of FVP from $2\times2$ to $256\times256$.}
\end{figure}

We also investigate different sizes of the FVP in Eq.~\eqref{eq:v_f_box}. 
We set the size $r$ from $2$ to $256$, 
and measure the Dice scores on the MM-WHS dataset with two adaptation directions.
The results in Fig.~\ref{fig:ablation_size} demonstrate that as the size of the prompt increases, 
the performance gradually improves, and then gradually decreases. 
It is probably because over-fitting occurs for large number of learnable parameters in FVP when the size is too big. 
The sizes $r=16$ and $r=32$ yield the best performance.

\begin{figure}[t]
    \centering
    \includegraphics[width=0.9\linewidth]{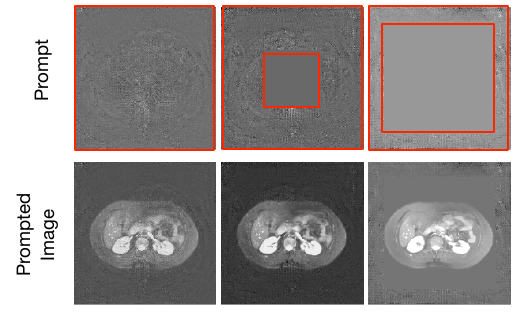}
    \caption{\label{fig:spatial}\textbf{ Visualization results for prompts in the spatial domain}. 
    The learned prompts are within the red bounding box. The prompt looks gray because of the negative values and the central part does not change. }
\end{figure}

\subsubsection{Comparison with the visual prompt in the spatial domain}

Different visual prompt designs in the spatial space were tested in~\cite{visualprompt} for transfer learning in image classification,  
including pixel patch in random and fixed locations, and padding. 
In this subsection, we would like to test the visual prompt (VP) $v$ parameterized in the spatial space~\cite{visualprompt} for SFUDA, 
called as spatial VP (SVP), 
and compare the results using SVP and FVP. 
For SVP, we use the padding design of SVP, 
considering experiments in~\cite{visualprompt} showed that the padding design obtains the best performance for image classification. 
Without loss of generality, denote a target image $x_t \in \mathbb{R}^{H \times W \times C}$ for the 2D case with spatial size $H\times W$ and channel $C$. 
Then, for the padding design of SVP with the padding parameter $s$, the number of trainable parameters of the VP is $2C(H-s)+2Cs(W-s)$, 
and the cuboid with the size of $(H-2s) \times (W-2s)$ at the center of SVP are zero. 
When $s=\max(H,W)$, the full FVP is learned without the zero cuboid. 
For the CT$\rightarrow$MRI task on the Abdominal dataset, we train the SVPs with three different padding parameters $s=256,156,56$, respectively, for a fair comparison. 
Then, the zero cuboids of SVPs have size of $0\times 0$, $100\times 100$, and $200\times 200$. 
The segmentation results of SFUDA are shown in Table~\ref{VP}, 
where FVP yields the best average Dice score compared with SVP with different padding parameters. 
It is mainly because FVP obtains the best performance on Spleen. 
Spleen is indeed the most difficult organ for this task, because the source-only method with the pre-trained model obtains a very low average Dice $0.009$. 
The learned prompts of SVP and three prompted images are shown in Fig.~\ref{fig:spatial}, 
where the prompted images of SVP are noisy with sharp boundary around the padding. %

\begin{table}    
  \caption{\textbf{Domain adaptation results of CT$\rightarrow$MRI task on the Abdominal dataset based on visual prompts in the spatial domain.} }
  
\label{VP}
\centering
\setlength{\tabcolsep}{3pt}
\begin{tabular}{cccccc}
\toprule
VP design & \textbf{Liver} & \textbf{R.kidney}  & \textbf{L.kidney} & \textbf{Spleen} &\textbf{Average} \\  
\hline
source-only & 0.582 & 0.789 & 0.686 & 0.009 & 0.517 \\
target supervised&0.727& 0.941&  0.941& 0.713&0.831 \\
\hline
  FVP ($r=32$)  &0.648 &0.876 & 0.803 & \textbf{0.605} &\textbf{0.733}\\
  SVP ($s=256$)  &0.708& \textbf{0.888}& 0.767& 0.092&0.614
 \\
  SVP ($s=156$) & \textbf{0.721}& 0.866&0.696& 0.138&0.605
 \\
SVP ($s=56$) &0.676& 0.862&\textbf{0.813}& 0.096& 0.612
\\
 \bottomrule
\end{tabular}
\end{table}


\begin{figure*}[t]
    \centering
    \includegraphics[width=0.9\linewidth]{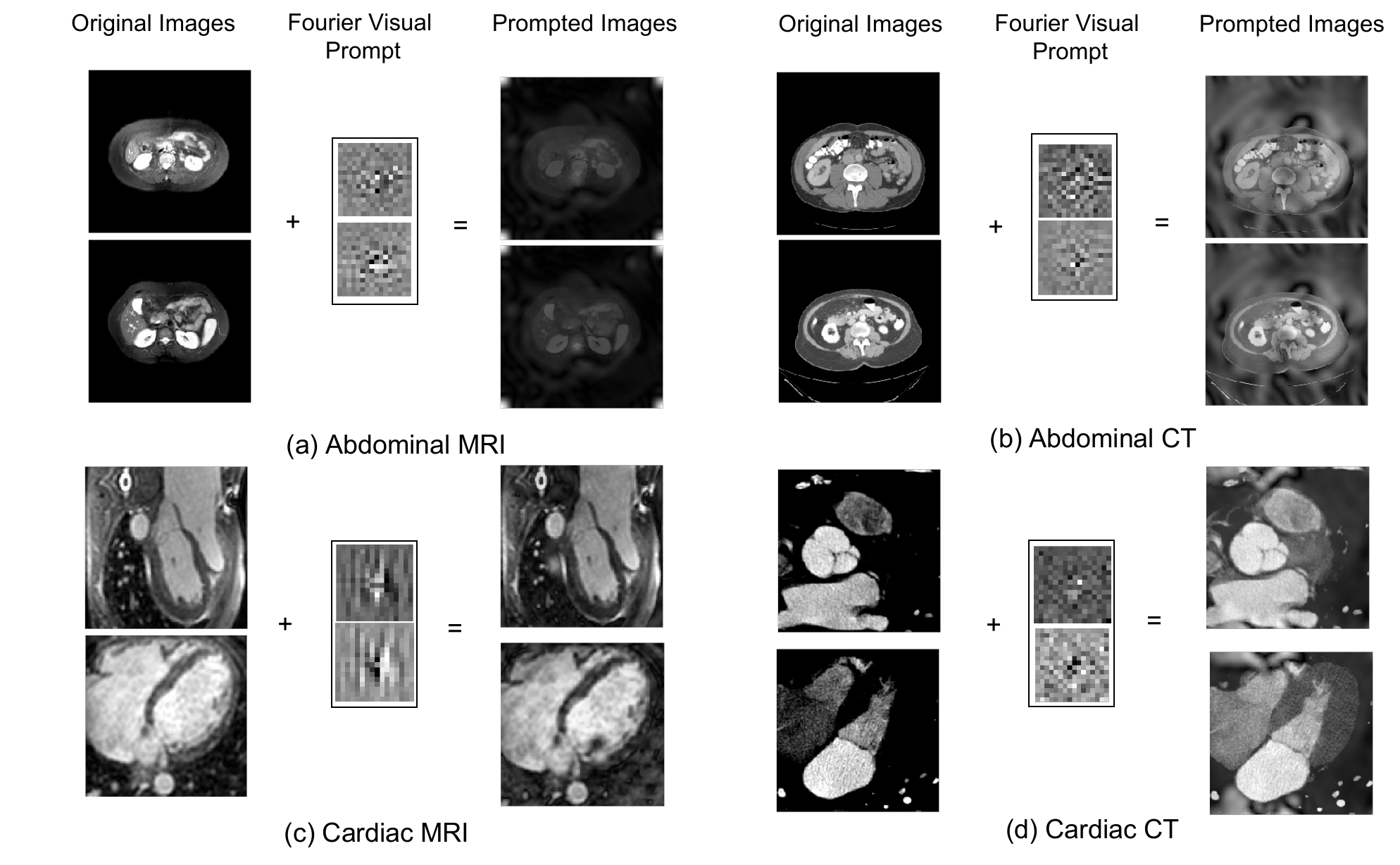}
    \caption{\label{fig:FVP_visual}
    \textbf{Visualization of the Fourier visual prompts (with real and imaginary parts), input images, and prompted images}. 
    (a) an MRI sample in CT$\rightarrow$MRI in Abdominal dataset. 
    (b) a CT sample in MRI$\rightarrow$CT in Abdominal dataset. 
    (c) an MRI sample in CT$\rightarrow$MRI in Cardiac dataset. 
    (d) a CT sample in MRI$\rightarrow$CT in Cardiac dataset.}
\end{figure*}

\subsubsection{Effect of Reliable Label Detection Module}

\begin{table}
  \caption{\label{tab:Multi-level}\textbf{Ablation studies of the reliable label detection module in CT$\rightarrow$MRI on the Abdominal dataset}.  
  Dice scores are listed for the method with or without global threshold, intra-class threshold, and prototype based selection.}
\centering
\setlength{\tabcolsep}{3pt}
\begin{tabular}{c|c|c|c|c}
\toprule
\multirow{2}{*}{\makecell{}{}{Method}}&\multicolumn{2}{c}{\makecell{Double-Threshold}{}{}}& \multirow{2}{*}{\makecell{}{}{Prototype}}&\multirow{2}{*}{\makecell{}{}{Dice}} \\
\cline{2-3}
& \makecell{}{}{global threshold}&\makecell{}{}{intra-class threshold}&\\
\hline \multirow{4}{*}{FVP} & &  &  &0.693\\ 
 &  &  &\checkmark& 0.710\\
& \checkmark &  & &0.702
 \\
 & \checkmark & \checkmark && 0.716
 \\

 & \checkmark & \checkmark &\checkmark&\textbf{0.733} \\
 \bottomrule
\end{tabular}

\end{table}  

In this subsection, we perform ablation studies for our proposed reliable label detection module in Section~\ref{sec:ST} in CT$\rightarrow$MRI on the Abdominal dataset. 
Table~\ref{tab:Multi-level} shows the Dice scores with or without global threshold, intra-class threshold, and prototype based selection.  
The experiment shows that our reliable label detection module with both double-threshold selection and prototype based selection yields the best Dice. 

\begin{figure}[t]
    \centering
    \includegraphics[width=1\linewidth]{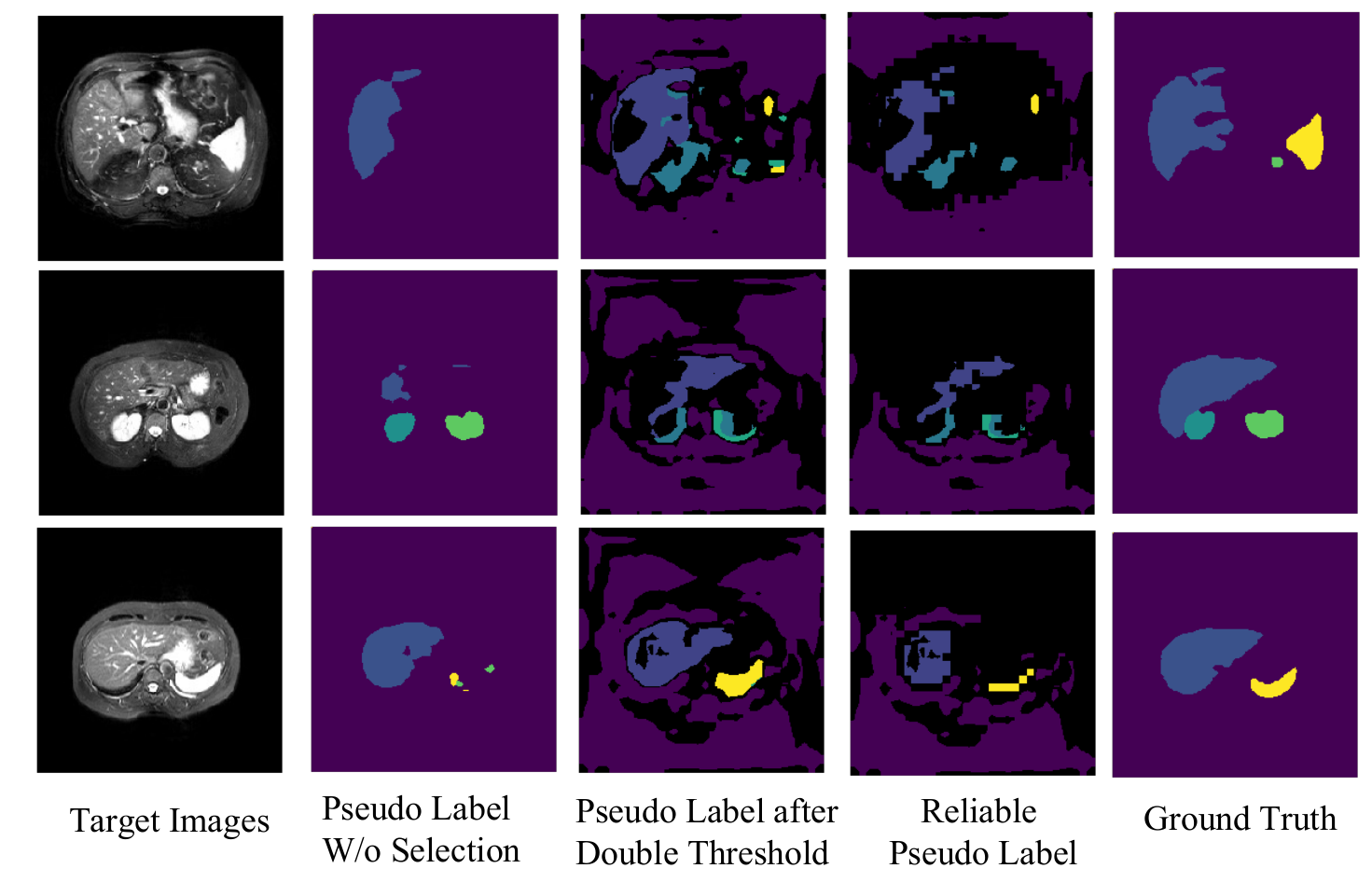}
    \caption{{\label{fig:DT}\textbf{Visualization of pseudo labels}. 
      From left to right: 
      Target images (MRI); 
      Pseudo labels from the pre-trained model (trained on CT); 
      Reliable pseudo labels using only the double-threshold selection (black means unreliable voxels); 
      The final reliable pseudo labels using both the double-threshold selection and the prototype-based selection; 
      The ground truth.} }
\end{figure}

The method without any pseudo label selections can achieve $0.693$ in Dice. 
FVP with only prototype-based selection or double-threshold selection can also alleviate the domain shift, but the performance is worse than FVP with both. 
It can be observed that all the components contribute to the final performance. 
We also provide three examples of reliable pseudo labels generated from the reliable label detection module, as shown in Fig.~\ref{fig:DT}. 
From the visualization results, we observe that the pseudo labels change after the double-threshold selection only, 
and after both selection methods in the reliable label detection module,  
especially for the voxels assigned as background. 
It demonstrates that the reliable label detection module could relieve the misprediction caused by the class imbalance issue of voxels.

\begin{figure}[t]
    \centering
    \includegraphics[width=1\linewidth]{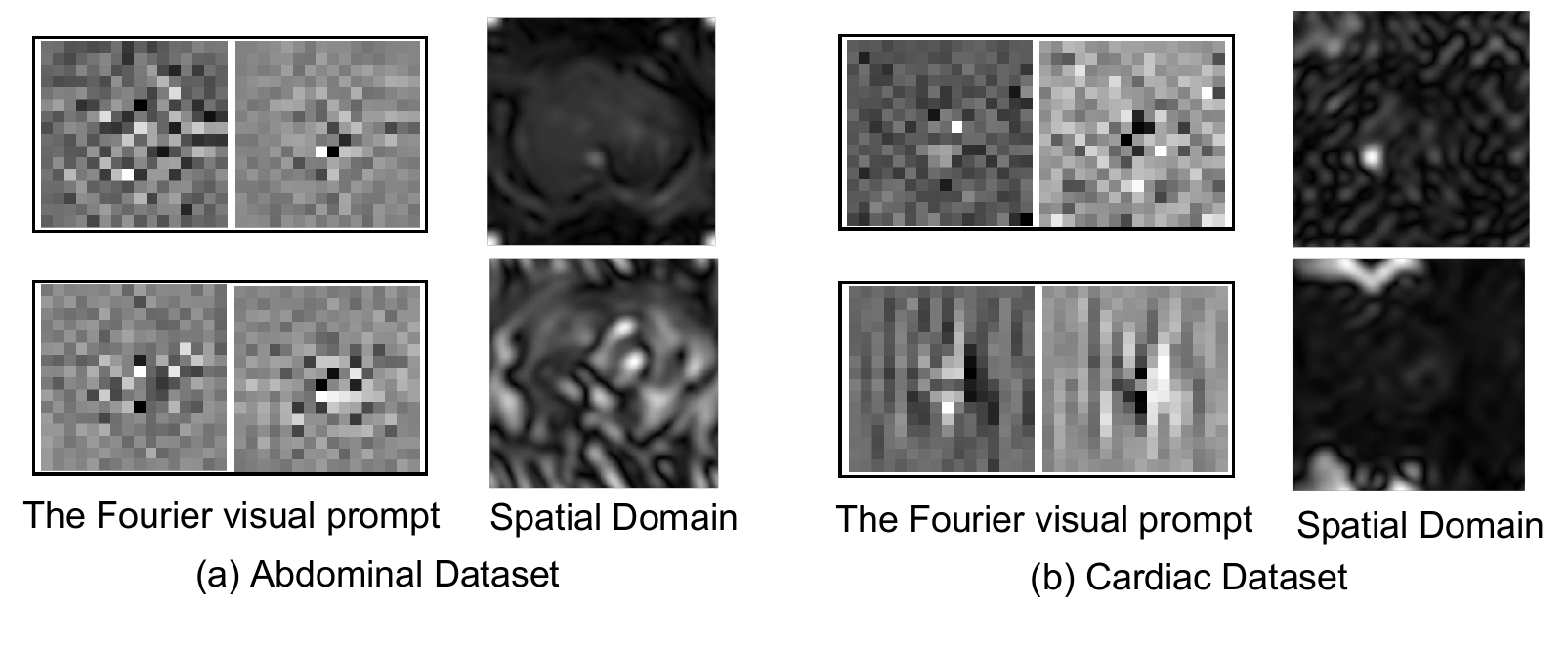}
    \caption{\label{fig:prompts}\textbf{Visualization for the learned Fourier visual prompts (with real and imaginary parts)}. 
    (a) and (b) are FVPs of two datasets. 
    The top row shows the FVPs in CT$\rightarrow$MRI, 
    and the bottom row shows the FVPs in MRI$\rightarrow$CT. 
    The size of prompts is $16 \times 16$ and the image size in the spatial domain is $256 \times 256$.}
\end{figure}
\begin{figure}[t]
    \centering
    \includegraphics[width=1\linewidth]{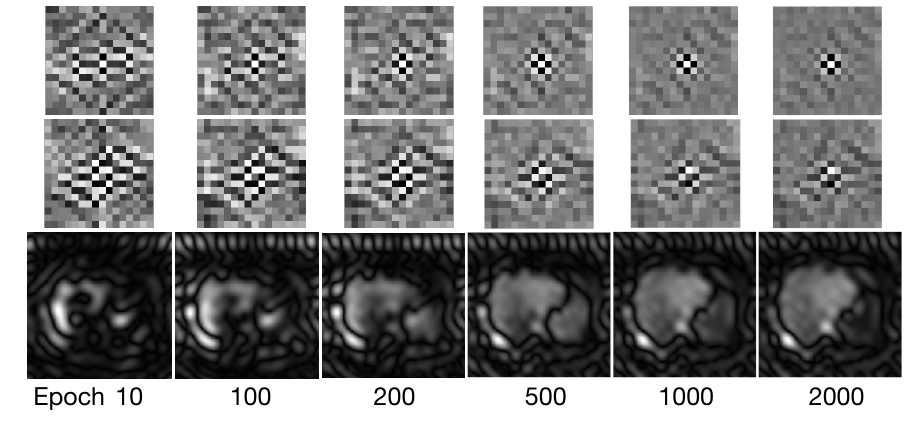}
    \caption{\label{fig:process}{ \textbf{Visualization of the training process of FVP.} 
    The first and second rows are visualization of the real and imaginary parts of the prompt in the frequency domain, respectively. 
    Note that the size of the prompt is $16\times16$. 
    The third row is the prompt in the spatial domain with a size of $256\times256$.
    }
    }
\end{figure}

\subsubsection{Performance on the U-Net backbone}

It should be noted that the proposed FVP essentially could work with various backbones as a plug-and-play SFUDA method. 
To validate this, we trained U-Net~\cite{Unet} on the source data as the pre-trained model, and then applied our proposed FVP to the U-Net backbone. 
The adaptation results of the U-Net backbone are shown in Table~\ref{backbone}, 
which demonstrates that our FVP also performs well with the U-Net backbone besides DeepLabV3.

\begin{table}
  \caption{\textbf{Domain adaptation results of CT$\rightarrow$MRI on the Abdominal dataset based on U-Net backbone.}}
\label{backbone}
\centering
\setlength{\tabcolsep}{3pt}
\begin{tabular}{cccccc}
\toprule                 
Method& \textbf{Liver} & \textbf{R.kidney}  & \textbf{L.kidney} & \textbf{Spleen} &\textbf{Average}\\  
\hline

source-only & 0.579 & 0.721 & 0.602 & 0.021 & 0.481
 \\
target supervised & 0.823 & 0.906 & 0.805 & 0.817 & 0.838
 \\
\hline
LD\cite{you2021domain}& 0.601 & 0.852 & 0.766 & 0.493 & 0.678\\
DPL\cite{chen2021source} & 0.588 & 0.843 & 0.779 & 0.482 & 0.673\\
OS\cite{liu2021adapting} &0.594 & 0.850 & \textbf{0.801} & 0.419 & 0.666\\
SFDA\cite{SFDA}&0.609&0.820&0.702&0.414&0.636\\
FSM\cite{FSM}&0.612&0.837&0.786&0.543&0.695\\
FVP (ours) &\textbf{0.643}&\textbf{0.866} &0.796 & \textbf{0.610} &\textbf{0.729}\\

 \bottomrule
\end{tabular}
\end{table}

\subsection{Visualization and Interpretability Analysis}

Fig.~\ref{fig:prompts} shows four groups of the Fourier visual prompts (FVPs) learned in our experiments with two datasets and two adaptation directions. 
The FVPs are shown in both the frequency space (with the size of $16 \times 16$) and the spatial voxel space. 
{FVP is learned using unlabeled target data with the frozen pre-trained model. 
Thus, the learned prompts are input-agnostic and domain-specific in the testing stage.}
Note that since the input image and the prompt image are both pre-processed with normalization of 0-mean and 1-variance as inputs of the pre-trained model, 
the center point of the FVP (indicating the mean) is not learnable. 
Thus, we set the center point of the FVP as the mean of surrounding points for visualization. 
Fig.~\ref{fig:FVP_visual} shows some examples of input images and prompted images with these learned FVPs. 
From Fig.~\ref{fig:prompts} and Fig.~\ref{fig:FVP_visual}, 
we can see that the prompted images by FVPs change the input images in a global manner. 
The prompted images are also consistent with the clinical interpretation. 
It is known that CT images are sensitive in bones, 
and MR images are sensitive in soft tissues. 

For example, the bone in abdominal CT images is bright, and it appears in low brightness after prompting, similarly as bone in MRI. 
For cardiac CT images, the contrast between the myocardium and other anatomical structures is high, 
and it decreases after prompting. 
Thus, a Fourier visual prompt could be used to change the visual style of the input image. %

{
We demonstrate the visualization of FVPs during the training process in Fig.~\ref{fig:process}. 
The first two rows in Fig.~\ref{fig:process} are the real and imaginary parts of prompts, 
and the last row is the prompts in the spatial domain after the inverse Fourier transform of FVP. 
We observe that during the training progress, the prompt changes more dramatically in the low-frequency part. }%

\section{Discussion}
{{%
As shown in Fig.~\ref{fig:framework}, 
compared with existing UDA and SFUDA methods~\cite{you2021domain,chen2021source,liu2021adapting, SFDA, FSM}, 
FVP is a novel method for SFUDA with the \emph{frozen} pre-trained model, 
where the visual prompt (VP) is parameterized using the low-frequency part of the input image in its frequency space. 
Thus, FVP has less number of trainable parameters and more efficient, compared with other SFUDA methods, as shown in Table~\ref{tab:number}. 
%
For the experiments of three dataset in Table~\ref{tab:Abdominal}, Table~\ref{tab:MMWHS} and Table~\ref{BraTS} with two adaptation directions, 
the proposed FVP yields the best Dice values in 5 cases over all 6 cases, and the best ASD values in all 6 cases. 
We think that the experiments clearly demonstrate that our FVP perform generally better compared with other DA methods. 
}}

In Table~\ref{tab:Abdominal}, the number of CT samples (30) is larger than the number of MRI samples (20). 
While the Dice score by source-only in CT$\rightarrow$MRI ($0.517$) is much lower than that in MRI$\rightarrow$CT ($0.647$). 
In Table~\ref{tab:MMWHS}, the Dice score by source-only in CT$\rightarrow$MRI ($0.412$) is much lower than that in MRI$\rightarrow$CT ($0.714$). 
Therefore, we think the adaptation of CT$\rightarrow$MRI is more difficult than MRI$\rightarrow$CT, 
which is consistent with discussions in~\cite{chen2020unsupervised}. 
We think it is probably because that MRI provides more texture details in organs, 
compared with CT, 
which makes training the source segmentation model in CT results in a worse model for adaptation than training in MRI. 

FVP does not work well on Dice in MRI$\rightarrow$CT adaptation in Table~\ref{tab:MMWHS}, 
probably because MRI$\rightarrow$CT is the least discriminating task among all four tasks in Table~\ref{tab:Abdominal} and~\ref{tab:MMWHS}. 
With the given six methods in Table~\ref{tab:Abdominal} and~\ref{tab:MMWHS}, 
for a specific task, we could define the Discrimination Score (DS) as the proportion by which the maximum Dice value exceeds the minimum Dice value by these six methods. 
Then, for Table~\ref{tab:Abdominal}, DS values are $17.7\%$ (CT$\rightarrow$MRI) and $24.4\%$ (MRI$\rightarrow$CT), respectively. 
For Table~\ref{tab:MMWHS}, DS values are $9.2\%$ (CT$\rightarrow$MRI) and $8.8\%$ (MRI$\rightarrow$CT), respectively. 
Thus, we argue that the smallest DS value ($8.8\%$) means MRI$\rightarrow$CT in Table~\ref{tab:MMWHS} is the least discriminating task among all four tasks. 
All six methods obtain similar Dice values with the smallest DS value for MRI$\rightarrow$CT in Table~\ref{tab:MMWHS}. 
Besides, the proposed FVP still yields the best ASD for MRI$\rightarrow$CT in Table~\ref{tab:MMWHS}, 
which demonstrates the effectiveness of FVP for this task. %

Although the proposed FVP works well for SFUDA with \emph{frozen} pre-trained models, 
here are some points which have not been considered in this paper and could be future work directions. 
First, 
we have not tested the proposed FVP with transformer-based backbones, 
since transformer normally requires more data for a good performance compared with CNN based backbones (e.g., DeepLabV3 and UNet). 
The main contribution of this paper is to propose FVP for SFUDA with the frozen pre-trained model, 
which is essentially a plug-and-play method working with arbitrary backbones. 
We have validated FVP with the DeepLabV3 backbone in Table~\ref{tab:Abdominal}, Table~\ref{tab:MMWHS} and Table~\ref{BraTS}, 
and with the U-Net~\cite{Unet} backbone in~\ref{backbone}.  
It could be a future work to pre-train a transformer based backbone using a larger number of data and apply FVP to the pre-trained transformer, 
but it is now outside of the scope of this paper. 
Second, 
if we not only learn the prompt, but also finetune the segmentation head of the pre-trained model, 
it may achieve a better segmentation result than FVP with the \emph{frozen} model.
However, since the purpose of this paper is to propose FVP for SFUDA with the \emph{frozen} pre-trained model (with both the backbone and the head) 
for better deployment in clinical scenario, 
we do not finetune the head in this paper. 
Third, the learned prompt is input-agnostic, while an input-specific prompt may have a better performance, which could be a future work.%

\section{Conclusion}

In the paper, we propose a novel method, called Fourier Visual Prompting (FVP), for Source-Free Unsupervised Domain Adaptation (SFUDA) of medical image segmentation.  
In FVP, a visual prompt is added to the input target image to steer the frozen pre-trained model to perform well in the target domain, 
without access to the source data and without requiring changing the pre-trained model.  
The visual prompt is parameterized using only a small amount of low-frequency learnable parameters in the input frequency space, 
so that it globally changes the input image. 
We also propose a reliable label detection module to learn the prompt, 
by minimizing the segmentation loss between the predicted segmentation of the prompted target image and reliable pseudo label of the target image under the frozen model. 
To our knowledge, FVP is the first work to apply visual prompt to SFUDA for medical image segmentation. 
The experiments in three public datasets demonstrate that the proposed FVP outperforms existing state-of-the-art SFUDA methods, 
indicating that leveraging the Fourier visual prompt could achieve domain adaptation in a simple yet effective way.  

\bibliographystyle{IEEEtran}
\bibliography{IEEEabrv.bib}

\end{document}